\documentclass{article}

\usepackage{microtype}
\usepackage{graphicx}
\usepackage{subfigure}
\usepackage{booktabs}
\usepackage{hyperref}
\usepackage[accepted]{icml2024}
\usepackage{amsmath}
\usepackage{amssymb}
\usepackage{float}
\usepackage{mathtools}
\usepackage{amsthm}
\usepackage[capitalize,noabbrev]{cleveref}
\theoremstyle{plain}

\theoremstyle{definition}

\theoremstyle{remark}

\usepackage[textsize=tiny]{todonotes}

\usepackage{amssymb}
\usepackage{color}
\usepackage{graphicx} 
\usepackage{multirow} 
\usepackage{booktabs} 
\usepackage{colortbl}
\usepackage{makecell}
\usepackage{bbding}
\usepackage{hyperref}
\usepackage{subfigure}
\usepackage{bbding}
\usepackage{titletoc}

\icmltitlerunning{EvTexture: Event-driven Texture Enhancement for Video Super-Resolution}

\newcommand{\etal}{et~al.}

\begin{document}
	
	\twocolumn[
	\icmltitle{EvTexture: Event-driven Texture Enhancement for Video Super-Resolution}
	
	
	
	\icmlsetsymbol{equal}{*}
	
	\begin{icmlauthorlist}
		\icmlauthor{Dachun Kai}{ustc}
		\icmlauthor{Jiayao Lu}{ustc}
		\icmlauthor{Yueyi Zhang}{ustc,iai}
		\icmlauthor{Xiaoyan Sun}{ustc,iai}
	\end{icmlauthorlist}
	
	\icmlaffiliation{ustc}{University of Science and Technology of China}
        \icmlaffiliation{iai}{Institute of Artificial Intelligence, Hefei Comprehensive National Science Center}
	
	\icmlcorrespondingauthor{Yueyi Zhang}{zhyuey@ustc.edu.cn}
	
	\icmlkeywords{Video Super-Resolution, Event Camera, Texture Enhancement, ICML}
	
	\vskip 0.2in
	]
	
	
	
	\printAffiliationsAndNotice{}  % leave blank if no need to mention equal contribution
	
	\begin{abstract}
	
     Event-based vision has drawn increasing attention due to its unique characteristics, such as high temporal resolution and high dynamic range. It has been used in video super-resolution (VSR) recently to enhance the flow estimation and temporal alignment. Rather than for motion learning, we propose in this paper the first VSR method that utilizes event signals for texture enhancement. Our method, called EvTexture, leverages high-frequency details of events to better recover texture regions in VSR. In our EvTexture, a new texture enhancement branch is presented. We further introduce an iterative texture enhancement module to progressively explore the high-temporal-resolution event information for texture restoration. This allows for gradual refinement of texture regions across multiple iterations, leading to more accurate and rich high-resolution details. Experimental results show that our EvTexture achieves state-of-the-art performance on four datasets. For the Vid4 dataset with rich textures, our method can get up to 4.67dB gain compared with recent event-based methods. Code: \url{https://github.com/DachunKai/EvTexture}.
     
	\end{abstract}

        \vspace{-0.6cm}
	\section{Introduction} 
	\label{sec:intro}
 
	\begin{figure}[t!]
		\centering
		\includegraphics[width=\columnwidth]{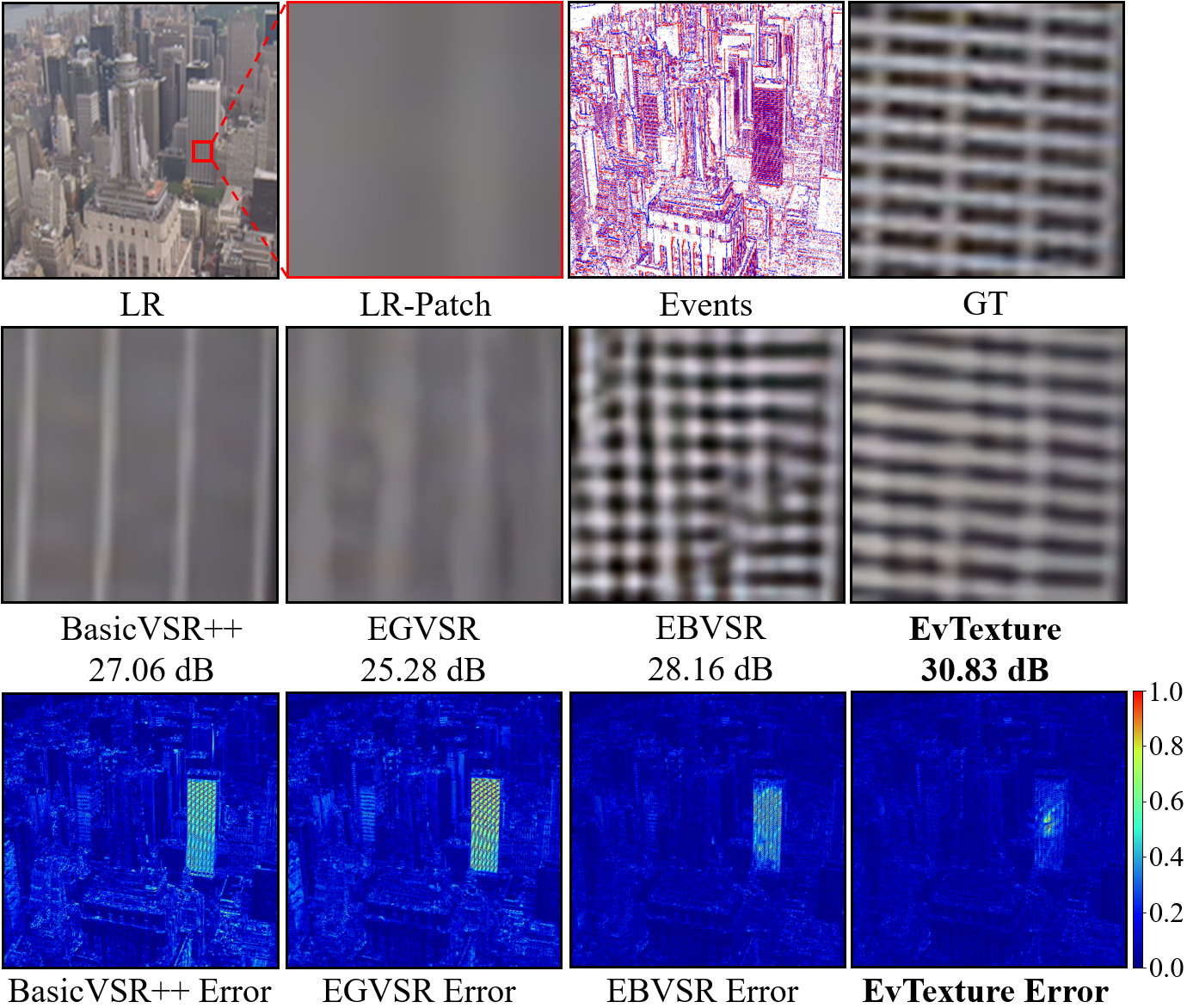}
        \vspace{-0.3cm}
		\caption{Comparative results of VSR methods on the City clip in Vid4~\cite{liu2013bayesian}. It can be observed that current VSR methods, with~\cite{lu2023learning,kai2023video} or without event signals~\cite{chan2022basicvsr++}, still suffer from blurry textures or jitter effects, resulting in large errors in texture regions. In contrast, our method can predict the texture regions successfully and greatly reduce errors in the restored frames.}
		\label{fig:fig1}
		\vspace{-0.2cm}
	\end{figure}
	
	\begin{figure*}[t!]
		\centering
		\vspace{0.1cm}
		\includegraphics[width=0.95\textwidth]{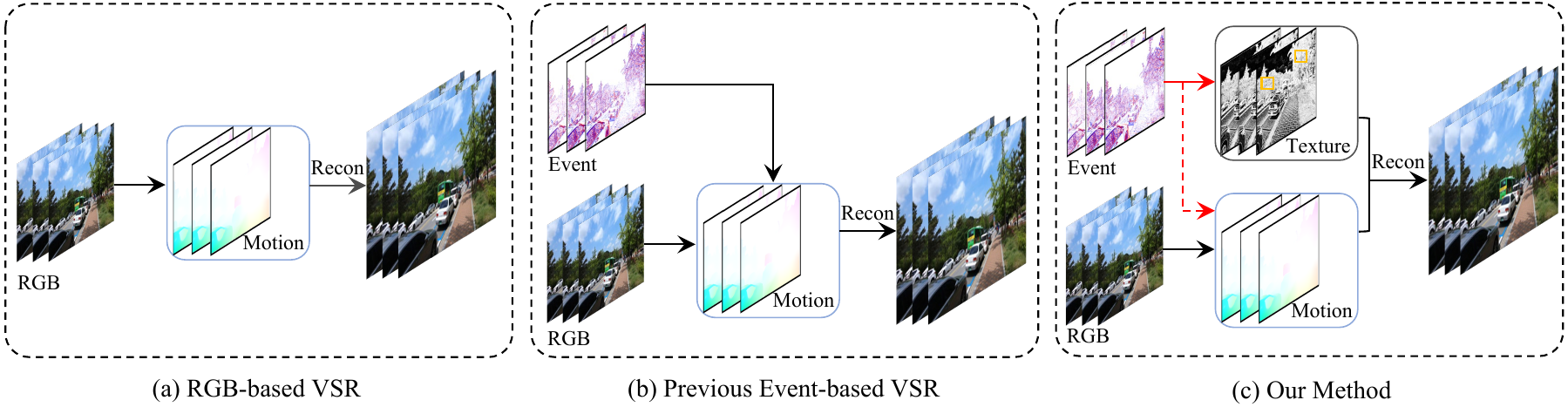}
        \vspace{-0.1cm}
		\caption{Comparisons of VSR learning process. RGB-based methods~\cite{chan2022basicvsr++,lin2022unsupervised} usually focus on motion leaning to recover the missing details from other unaligned frames. Previous event-based methods~\cite{lu2023learning,kai2023video} have attempted to use events to enhance the motion learning. In contrast, our method is the first to utilize events to enhance the texture restoration in VSR. The red dotted line is an optional branch, where our method can easily adapt to approaches that use events to enhance the motion learning.}
		\label{fig:fig2}
		\vspace{-0.2cm}
	\end{figure*}
	
    Video super-resolution (VSR) aims at restoring high-resolution (HR) videos from their low-resolution (LR) counterparts. It has extensive applications in various domains such as surveillance~\cite{zhang2010super}, virtual reality~\cite{liu2020single} and video enhancement~\cite{xue2019video}. Compared to single image super-resolution, VSR pays more attention to modeling the temporal relationships between frames, as it tries to predict missing details of the current HR frame from other unaligned frames.

    Recently, event signals captured by event cameras~\cite{lichtsteiner2008128} have been used in VSR~\cite{jing2021turning,lu2023learning,kai2023video}. Compared with standard cameras, event cameras have very high temporal resolution as well as high dynamic range~\cite{gallego2020event}. They thus can provide complementary motion information for VSR. For instance, EGVSR~\cite{lu2023learning} introduces a temporal filter branch to explore the motion information from events, such as edges and corners. EBVSR~\cite{kai2023video} utilizes events to enhance the flow estimation and temporal alignment in VSR. However, as shown in Fig.~\ref{fig:fig1}, these methods still suffer from large errors in texture regions.

    Texture restoration is a very challenging problem in VSR. It is difficult to restore HR textural details from the corresponding LR ones. We notice that there are some attempts working on texture enhancement for image super-resolution~\cite{cai2022tdpn}, while for VSR, most methods focus on addressing issues caused by large motion~\cite{wang2019edvr,lin2022unsupervised} or occlusions~\cite{chan2022basicvsr++}. To our best knowledge, little work has been presented on texture restoration in VSR so far.
	
    We observe that event signals are not only with high temporal resolution but also full of high-frequency dynamic details, which are desirable for texture restoration in VSR. We thus propose utilizing the event signal for texture restoration and present an event-driven texture enhancement neural network, EvTexture, for VSR. Unlike other event-related methods that use event signals directly in HR frame estimation, our EvTexture progressively recovers the high-frequency textural information in two ways. Firstly, we present a two-branch structure in which the texture enhancement branch is introduced in addition to the motion branch to enhance the texture details. Secondly, we present an Iterative Texture Enhancement (ITE) module to progressively explore the high-temporal-resolution event information for texture restoration. This approach enables a step-by-step enhancement of texture regions across multiple iterations, resulting in more accurate and rich HR details. Experimental results on four datasets demonstrate the effectiveness of our proposed EvTexture. Our EvTexture significantly enhances the performance of VSR especially in texture regions.
	
	The main contributions of this paper are as follows:
	\begin{itemize}
            \vspace{-1ex}
		\item We propose the first event-driven scheme for texture restoration in VSR.
            \vspace{-1ex}
		\item We propose recovering high-frequency textural information progressively by our presented texture enhancement branch coupled with an ITE module.
            \vspace{-1ex}
		\item Our proposed texture restoration method achieves state-of-the-art performance on four VSR benchmarks and especially excels in restoring texture-rich clips.
	\end{itemize}
        
	\section{Related Work}
	
	\begin{figure*}[t!]
		\centering
		\vspace{0.1cm}
		\includegraphics[width=0.98\textwidth]{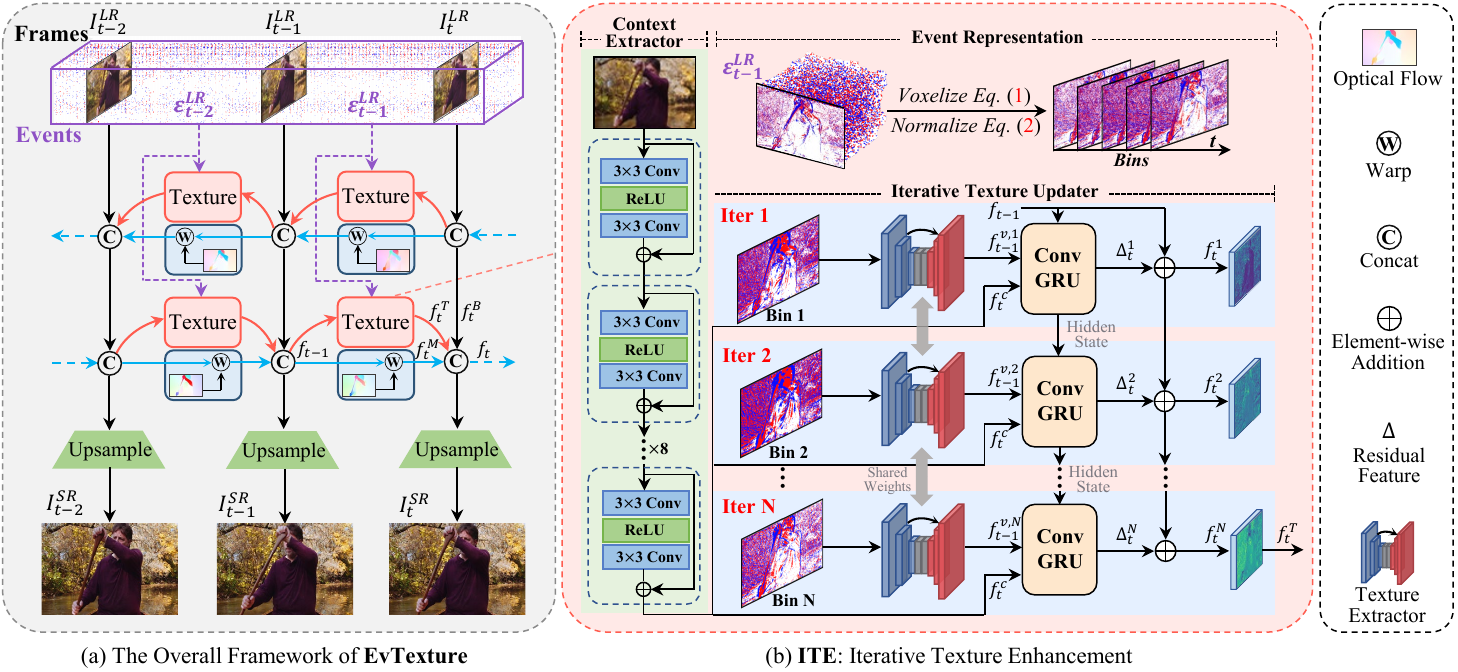}
		\caption{Network architecture of EvTexture. (a) EvTexture adopts a bidirectional recurrent network, where features are propagated forward and backward. At each timestamp, it includes a motion branch and a parallel texture branch to explicitly enhance the restoration of texture regions. (b) In the texture branch, the ITE module plays a key role. It progressively refines the feature across multiple iterations, leveraging high-frequency textural information from events along with context information from the current frame.}
		\label{fig:fig3}
		\vspace{-0.2cm}
	\end{figure*}
	
    \subsection{Video Super-Resolution}
    Based on whether the input data uses solely RGB frames or includes additional event signals, VSR methods can be categorized as RGB-based VSR or event-based VSR.
	
    \paragraph{RGB-based VSR.}  Previous RGB-based VSR methods primarily exploit temporal redundancy between neighboring frames, emphasizing motion learning (Fig.~\ref{fig:fig2}(a)). Techniques like optical flow estimation~\cite{xiao2021space,chan2021basicvsr,chan2022basicvsr++} and deformable convolutions~\cite{wang2019edvr,liang2022recurrent} are used to create advanced alignment modules. Typically, Lin~\etal~\yrcite{lin2022unsupervised} proposed an unsupervised flow-aligned sequence-to-sequence model (S2SVR) to model the inter-frame relation. However, while these methods get good results in large motion regions, they struggle for hard cases with complex textures, as high-frequency information is lost in LR videos.
	
    \vspace*{-1ex}
    \paragraph{Event-based VSR.}  Event cameras, also known as neuromorphic cameras, are a new type of visual sensor that operates on the principle of asynchronously capturing brightness changes~\cite{gallego2020event}. Incorporating event signals into VSR tasks has gained great attention~\cite{jing2021turning,lu2023learning,kai2023video}. These techniques mainly focus on using events to enhance the motion learning (Fig.~\ref{fig:fig2}(b)). Recently, Kai~\etal~\yrcite{kai2023video} proposed to estimate nonlinear flow from events to enhance the temporal alignment in VSR. However, they have not fully exploited the high-frequency dynamic details of events to address the significant challenges associated with texture regions.
	
    \vspace*{-1ex}
    \subsection{Texture Restoration}
    We study the fundamental issue of texture restoration in VSR. This challenging problem not only needs to recover high-frequency details but also ensures temporal consistency during video playback given the rich details. While there have been efforts in texture region enhancement in single image super-resolution~\cite{ma2021structure,cai2022tdpn}, little work has been dedicated to texture restoration in VSR.
	
    \vspace*{-1ex}
    \subsection{Iterative Refinement}
    Recently, Teed \& Deng~\yrcite{teed2020raft} proposed to iteratively update a flow field, using a recurrent GRU-based update operator. This idea has gradually been applied to other tasks as well, such as stereo~\cite{lipson2021raft}, monocular~\cite{shao2023IEBins}, and event-based flow estimation~\cite{wan2022learning}. For instance, Wan~\etal~\yrcite{wan2022learning} proposed an iterative update network structure to estimate temporally continuous and spatially dense flow by utilizing frame-event two modalities. In this paper, we take advantage of high-temporal-resolution event signals and employ a GRU-based iterative optimizer to improve the restoration of texture regions in VSR.
	
    \section{Preliminary}
	
    \subsection{Event Representation}
    The original event stream can be denoted as a set of 4-tuples $\mathcal{E} = \{e_k\}_{k=1}^{N_e}$, where $N_e$ represents the number of events. Each event $e_k$ contains four attributes: $ (x_k,y_k,t_k,p_k)$, where $(x_k,y_k)$, $t_k$ and $p_k$ represent the spatial coordinate, timestamp, and polarity of brightness change, respectively.
	
    In practice, it is challenging to find an appropriate representation for an event stream that preserves complete temporal information. In this work, we represent the event stream as a grid-like event voxel grid $\mathcal{V}$ as in the previous work~\cite{zhu2019unsupervised}, which discretizes the time domain into $B$ time bins. Each time bin in the voxel grid is described as:
	\begin{equation}\label{eq1}
		\mathcal{V}(i)=\sum_k p_k \max \left(0,1-\left|\left(i-1\right)-\frac{t_k-t_0}{t_{N_e}-t_{0}}\left(B-1\right)\right|\right),
	\end{equation}
	where $i \in\{1, \cdots, B\}$ represents the $i$-th time bin. In our experiments, consistent with previous studies~\cite{weng2021event,wan2022learning}, we also set $B=5$. Furthermore, to mitigate the impact of hot pixels, we follow the study~\cite{zhu2021eventgan} and normalize the voxel grid $\mathcal{V}$ as:
	\begin{equation}\label{eq2}
		\hat{\mathcal{V}}(i)=\frac{\min \left(\mathcal{V}(i), \eta\right)}{\eta},
	\end{equation}
	where $\eta$ is the 98th percentile value in the non-zero values of $\mathcal{V}$. In this way, we obtain the normalized voxel grid $\hat{\mathcal{V}}\in\mathbb{R}^{H \times W \times B}$, which contains rich high-frequency textural information (``Events'' of Fig.~\ref{fig:fig1}) and is easily processed by current deep neural networks.

    \section{Methodology}
	
    We propose a novel neural network, named \textbf{EvTexture}, to overcome the challenge of texture restoration in VSR by leveraging high-frequency event signals. The architecture of EvTexture is shown in Fig.~\ref{fig:fig3}. The input is an LR image sequence $[I^{LR}_t]_{t=1}^{T}$ consisting of $T$ frames, and the inter-frame events of the $T-1$ intervals $[\mathcal{E}^{LR}_t]_{t=1}^{T-1}$. The output is the corresponding super-resolved image sequence $[I^{SR}_t]_{t=1}^{T}$. 
	
    Our EvTexture adopts a bidirectional recurrent structure, based on BasicVSR~\cite{chan2021basicvsr}, where features are propagated forward and backward, and propagation modules are interconnected. At each timestamp, it employs a two-branch structure: a motion learning branch and a parallel texture enhancement branch. The former utilizes optical flow to align frames, while the latter leverages events to enhance the texture details. Then, features from both branches are fused and propagated to the next timestamp. Finally, the output features are upsampled through pixel shuffle~\cite{shi2016real} layers to reconstruct the HR frames.
	
    \subsection{Two-branch Structure}
	
    We illustrate the feature learning process in two branches using forward propagation as an example. The only difference in backward propagation is the direction of data flow. 
	
    \paragraph{Motion Learning Branch.}  Within the motion learning branch (the blue box of Fig. \ref{fig:fig3}(a)), following BasicVSR, we use a lightweight optical flow estimation network $\mathcal{S}$ (SpyNet)~\cite{ranjan2017optical} to estimate the optical flow and use it to align the propagation feature. Specifically, at a given timestamp $t$, we first take $I^{LR}_{t}$ and $I^{LR}_{t-1}$ as inputs for $\mathcal{S}$, and estimate the flow $F_{t\to t-1}$. We then use this flow to warp the feature $f_{t-1}$ to the current timestamp $t$, obtaining the feature $f_{t}^{M}$. This process can be expressed as:
    \begin{equation}\label{eq3}
		F_{t\to t-1}=\mathcal{S}\left(I_t^{L R}, I_{t-1}^{L R}\right), \; f_{t}^{M}=\mathcal{W}\left(f_{t-1}, F_{t\to t-1}\right),
    \end{equation}
    where $\mathcal{W}(\cdot)$ denotes the spatial warping operation. The motion branch provides basic features that can restore most of the simple and smooth regions.
	
    \paragraph{Texture Enhancement Branch.} Due to the rich high-frequency details in texture regions, a flow-based alignment method may not handle them adequately, and in some cases, it can even be detrimental~\cite{shi2022rethinking}. To address this challenge, we introduce an additional texture enhancement branch to enhance the texture regions explicitly, with the aid of event signals. Specifically, given the event stream $\mathcal{E}^{LR}_{t-1}$ between $I^{LR}_{t}$ and $I^{LR}_{t-1}$, the texture branch propagates $f_{t-1}$ and get the texture-enhanced feature $f^{T}_{t}$ as:
	\begin{equation}\label{eq4}
		f^{T}_{t}=\mathcal{A}\left(f_{t-1}, \mathcal{E}^{LR}_{t-1}, I^{LR}_{t} \right),
	\end{equation}
    where $\mathcal{A}(\cdot)$ refers to our novel ITE module that will be introduced in the following subsection. The texture branch greatly enhances the restoration of texture regions by leveraging high-frequency details from event signals.
	
    Our method adopts an iterative optimization design to optimize textures over time, with well-suited voxelizing of event streams into different temporal segments, as depicted in Eqs.~(\ref{eq1}) and~(\ref{eq2}). With this design, we progressively transfer texture details from events to the propagation feature, thus enhancing the restoration of complex texture regions.
	
    \subsection{Iterative Texture Enhancement}\label{sec:sec3.3}
	
    Directly incorporating event signals into the reconstruction of HR frames neglects to explore the potential relationship between the high-temporal-resolution properties of events and the high-frequency texture regions. In this work, motivated by an iterative structure~\cite{teed2020raft}, we propose the ITE module to better model the temporal relation between voxel bins and enhance the texture regions. The architecture of this module is shown in Fig.~\ref{fig:fig3}(b), which comprises two feature extractors and a GRU-based iterative texture updater.

    \paragraph{Feature Extractors.} We employ two types of feature extractors: the context extractor $\mathcal{C}$ and the texture extractor $\mathcal{T}$. The parameters of these extractors are shared across all iterations. The context extractor consists of eight residual blocks used in~\cite{wang2018esrgan} and is applied to the current frame $I_t^{LR}$ to extract the context feature $f^c_t$. The texture extractor aims to extract the texture feature $f^{v,i}_{t-1}$ from a voxel bin $\hat{\mathcal{V}}_{t-1}(i)$ during each iteration. It is implemented using a custom five-layer UNet~\cite{ronneberger2015u}, inspired by its robust ability to capture spatial-temporal features~\cite{jiang2018super}. The feature extraction process can be formulated as:
	
    \begin{equation}\label{eq5}
        f^c_t=\mathcal{C}\left(I_t^{L R}\right), {f}^{v,i}_{t-1}=\mathcal{T}\left(\hat{\mathcal{V}}_{t-1}(i)\right).
    \end{equation}
	
    Here, both the context feature $f^c_t$ and texture feature $f^{v,i}_{t-1}$ have a common feature size of $ \mathbb{R}^{C\times H \times W}$, matching the shape of the propagation feature $f_{t-1}$.
    
        \begin{table*}[t]
    \caption{Quantitative comparison (PSNR$\uparrow$/SSIM$\uparrow$) on Vid4~\cite{liu2013bayesian}, REDS4~\cite{nah2019ntire} and Vimeo-90K-T~\cite{xue2019video}  for 4$\times$ VSR. All results are calculated on Y-channel except REDS4 (RGB-channel). The input types ``I" and ``I+E'' represent RGB-based and event-based methods, respectively. \textcolor{red}{\textbf{Red}} and \textcolor{blue}{\underline{blue}} colors indicate the best and second-best performances, respectively.}
    \label{table:table1}
    \vskip 0.15in
    \centering
    \resizebox{\textwidth}{!}{
        \begin{tabular}{l|c|ccccccc}
            \toprule
            \multirow{2}[2]{*}{Method} & \multirow{2}[2]{*}{\makecell{Input \\ Type}} &  \multicolumn{5}{c}{Vid4} & \multirow{2}[2]{*}{REDS4} & \multirow{2}[2]{*}{Vimeo-90K-T} \\[-0.1em]
            
            \cmidrule(lr){3-7} 
            & & Calendar & City & Foliage & Walk & Average & &  \\
            
            \midrule
            EDVR~\cite{wang2019edvr} & I & 23.98/0.8143 & 27.83/0.8112 & 26.34/0.7560  & 31.06/0.9153 & 27.30/0.8242 & 31.09/0.8800 & 37.61/0.9489  \\
            BasicVSR~\cite{chan2021basicvsr} & I & 23.87/0.8094 & 27.66/0.8050 & 26.47/0.7710 & 30.96/0.9148 & 27.32/0.8265 & 31.42/0.8909 & 37.18/0.9450 \\
            IconVSR~\cite{chan2021basicvsr} & I & 24.07/0.8143  & 27.86/0.8111 & 26.54/0.7705 & 31.08/0.9158 & 27.46/0.8290 & 31.67/0.8948 & 37.47/0.9476  \\
            RTVAR~\cite{zhou2022revisiting} & I &24.65/0.8270 & 29.92/0.8428 & 26.41/0.7652 & 31.15/0.9167 & 27.90/0.8380  & 31.30/0.8850 & 37.84/0.9498 \\
            BasicVSR++~\cite{chan2022basicvsr++} & I &24.50/0.8288 & 28.05/0.8212 & 26.90/0.7868 & 31.71/0.9236 & 27.87/0.8413 &32.39/0.9069 & 37.79/0.9500 \\
            RVRT~\cite{liang2022recurrent} & I &24.55/0.8334 & 28.35/0.8363 & 26.98/0.7824 & 31.86/0.9251 & 27.94/0.8443 & 32.75/0.9113 & 38.15/0.9527 \\
            VRT~\cite{liang2022vrt} & I &24.52/0.8296 & 28.33/0.8308 & 26.78/0.7754 & 31.89/0.9258 & 27.88/0.8404 & 32.19/0.9006 & 38.20/0.9530 \\
            EGVSR~\cite{lu2023learning} & I+E  &21.53/0.6932 & 26.01/0.7068 & 24.33/0.6651 & 27.39/0.8574 & 24.84/0.7330 & 26.87/0.7790 & 34.62/0.9185 \\
            EBVSR~\cite{kai2023video} & I+E & 25.17/0.8548 & 29.30/0.8846 & 27.31/0.8187 & 31.91/0.9265 & 28.46/0.8701 & 31.47/0.8919 & 37.56/0.9490 \\
            \midrule
            \textbf{EvTexture} & I+E  &\textcolor{blue}{\underline{26.10}}/\textcolor{blue}{\underline{0.8756}} & \textcolor{blue}{\underline{31.24}}/\textcolor{blue}{\underline{0.9087}} & \textcolor{blue}{\underline{28.12}}/\textcolor{blue}{\underline{0.8475}} & \textcolor{blue}{\underline{32.67}}/\textcolor{blue}{\underline{0.9366}} & \textcolor{blue}{\underline{29.51}}/\textcolor{blue}{\underline{0.8909}} & \textcolor{blue}{\underline{32.79}}/\textcolor{blue}{\underline{0.9174}} & \textcolor{blue}{\underline{38.23}}/\textcolor{blue}{\underline{0.9544}}\\
            \textbf{EvTexture+} & I+E  &\textcolor{red}{\textbf{26.44}}/\textcolor{red}{\textbf{0.8859}} & \textcolor{red}{\textbf{31.82}}/\textcolor{red}{\textbf{0.9217}} & \textcolor{red}{\textbf{28.21}}/\textcolor{red}{\textbf{0.8542}} & \textcolor{red}{\textbf{32.86}}/\textcolor{red}{\textbf{0.9381}} & \textcolor{red}{\textbf{29.78}}/\textcolor{red}{\textbf{0.8983}} & \textcolor{red}{\textbf{32.93}}/\textcolor{red}{\textbf{0.9195}} &  \textcolor{red}{\textbf{38.32}}/\textcolor{red}{\textbf{0.9558}}\\
            \bottomrule
        \end{tabular}
    }
    \vskip -0.2in
    \end{table*}

    \paragraph{Iterative Texture Updater.} After extracting context and texture features, we introduce our texture updater, designed to transfer the textural information from each voxel bin and the context information from the current frame to the propagation feature. The texture updater, shared across each iteration, consists of three ConvGRU~\cite{ballas2015delving} layers and five residual blocks~\cite{wang2018esrgan}, denoted as $\mathcal{G}$ and $\mathcal{R}$, respectively. The propagation feature $f_t^i$, initialized with $f_{t-1}$, is updated in a residual manner as:
    \begin{equation}\label{eq6}
        \begin{aligned}
            & h_t^i = \mathcal{G}\left(h^{i-1}_t,\left[f^c_t,f^{v,i}_{t-1}\right]\right), \\ & \Delta^i_t = \mathcal{R}\left(h^i_t\right), \;
            f_t^i = f_t^{i-1} + \Delta^i_t .
        \end{aligned}
    \end{equation}
    Here, $h^i_t$ is the hidden state in each iteration at timestamp $t$, also initialized with $f_{t-1}$. The superscript $i$ represents the $i$-th iteration, where $i \in\{1, \cdots, N\}$. The iteration number $N$ is equal to the number of voxel bins, namely $B$. $[\cdot, \cdot]$ is the concatenating operation. After $N$ iterations, we obtain the enhanced texture feature $f_t^{T}$ as:
    \begin{equation}\label{eq7}
        f_t^{T} = f_{t-1} + \sum\nolimits_{i=1}^{N} \Delta^i_t,
    \end{equation}
    which includes the rich textural details transferred from each voxel bin. Our ITE module efficiently transfers high-frequency textual information from event signals to the propagation features and enables progressive enhancement of textural details across multiple iterations.

        \begin{table}[t]
    \caption{Comparison of perceptual similarity (LPIPS$\downarrow$), parameters and runtime on Vid4 and REDS4 for 4$\times$ VSR. The average runtime is computed for a clip containing 10 frames, each with an LR frame size of 180$\times$320, on an NVIDIA V100-16GB GPU.}
    \label{table:table2}
    \vskip 0.15in
    \centering
    \resizebox{\columnwidth}{!}{
        \begin{tabular}{ l|cccc}
            \toprule
            \multirow{2}[1]{*}{\makecell{Method}} &  \multirow{2}[1]{*}{\makecell{Vid4}} & \multirow{2}[1]{*}{\makecell{REDS4}}   & \multirow{2}[1]{*}{\makecell{\#Params \\ (M)}}  & \multirow{2}[1]{*}{\makecell{Runtime \\ (ms)}} \\[1.4em]
            \midrule
            EDVR~\cite{wang2019edvr} & 0.2641 & 0.2091  & 20.6 & 378 \\
            BasicVSR~\cite{chan2021basicvsr} & 0.2783 & 0.2018 & 6.3 & \textbf{63} \\
            IconVSR~\cite{chan2021basicvsr} & 0.2722 & 0.1946 & 8.7 & 70 \\
            BasicVSR++~\cite{chan2022basicvsr++} & 0.2593 & 0.1786   & 7.3 & 77 \\
            RVRT~\cite{liang2022recurrent} & 0.2481 & 0.1728 & 10.8 & 183 \\
            VRT~\cite{liang2022vrt} & 0.2505 & 0.1864 & 35.6 & 243 \\
            EGVSR~\cite{lu2023learning} & 0.3351 & 0.3024 & \textbf{2.6} & 193 \\
            EBVSR~\cite{kai2023video} & 0.2476 & 0.1996 & 12.2 & 92 \\
            \midrule
            \textbf{EvTexture} & \textcolor{blue}{\underline{0.2185}} & \textcolor{blue}{\underline{0.1684}} & 8.9 & 136 \\
            \textbf{EvTexture+} & \textcolor{red}{\textbf{0.2048}} & \textcolor{red}{\textbf{0.1642}}   & 10.1 & 139\\
            \bottomrule
        \end{tabular}
    }
    \vskip -0.1in
    \end{table}
        \begin{table}[t]
    \caption{Quantitative results on CED~\cite{scheerlinck2019ced} for \textbf{$2\times$} and \textbf{$4\times$} VSR. Metrics are calculated on the RGB-channel. $^\dagger$ denotes results are from EGVSR~\cite{lu2023learning}.}
    \label{table:table3}
    \vskip 0.15in
    \centering
    \resizebox*{\linewidth}{!}{
        \begin{tabular}{l|c|cccc}
            \toprule
            \multirow{2}[1]{*}{Method} & \multirow{2}[2]{*}{\makecell{Input \\ Type}} & \multicolumn{2}{c}{$2\times$} & \multicolumn{2}{c}{$4\times$} \\ [-0.1em]
            \cmidrule(lr){3-4} \cmidrule(lr){5-6}
            & & PSNR & SSIM & PSNR & SSIM \\
            \midrule
            DUF~\cite{jo2018deep}$^\dagger$ & I & 31.09 & 0.9183 & 24.43 & 0.8177 \\
            SOF~\cite{wang2020deep}$^\dagger$ & I & 31.84 & 0.9226 & 27.00 & 0.8050 \\
            TDAN~\cite{tian2020tdan}$^\dagger$ & I & 33.74 & 0.9398 & 27.88 & 0.8231\\
            RBPN~\cite{haris2019recurrent}$^\dagger$ & I & 36.66 & 0.9754 & 29.80 & 0.8975\\
            BasicVSR~\cite{chan2021basicvsr} & I & 39.57 & 0.9778 & 32.93 & 0.9001\\
            E-VSR~\cite{jing2021turning}$^\dagger$ & I+E & 37.32 & 0.9783 & 30.15 & 0.9053\\
            EGVSR~\cite{lu2023learning}$^\dagger$ & I+E & 38.69 & 0.9771 & 31.12 & \textcolor{red}{\textbf{0.9211}}\\
            EBVSR~\cite{kai2023video} & I+E & 40.14 & 0.9801 & 33.42 & 0.9075\\
            \midrule
            \textbf{EvTexture} & I+E & \textcolor{blue}{\underline{40.52}} & \textcolor{blue}{\underline{0.9813}} & \textcolor{blue}{\underline{33.68}} & 0.9112\\
            \textbf{EvTexture}+ & I+E & \textcolor{red}{\textbf{40.57}} & \textcolor{red}{\textbf{0.9815}} & \textcolor{red}{\textbf{33.71}} & \textcolor{blue}{\underline{0.9126}}\\
            \bottomrule
        \end{tabular}
    }
    \vskip -0.1in
    \end{table}

    \subsection{Feature Fusion}
	
    After the two-branch feature learning, we obtain the motion feature $f_{t}^{M}$ and the texture-enhanced feature $f_{t}^{T}$, respectively. We then employ an effective fusion design to aggregate motion and texture features and generate the propagation feature $f_t$ at timestamp $t$ as:
    \begin{equation}\label{eq8}
        f_t = \mathcal{R}\left(I_t^{LR},f_{t}^{B},\left[f_t^M,f_t^T\right]\right).
    \end{equation}
    Here, $f^B_t$ is the feature from the backward branch. Finally, the fused feature $f_t$ is passed through pixel shuffle layers for upsampling. The upsampled feature is then added with the bicubic upsampled result of the input frame $I_t^{LR}$ through element-wise addition to obtain the SR frame $I_t^{SR}$.

     \begin{figure*}[t!]
        \centering
        \vspace{0.1cm}
        \includegraphics[width=0.9\textwidth]{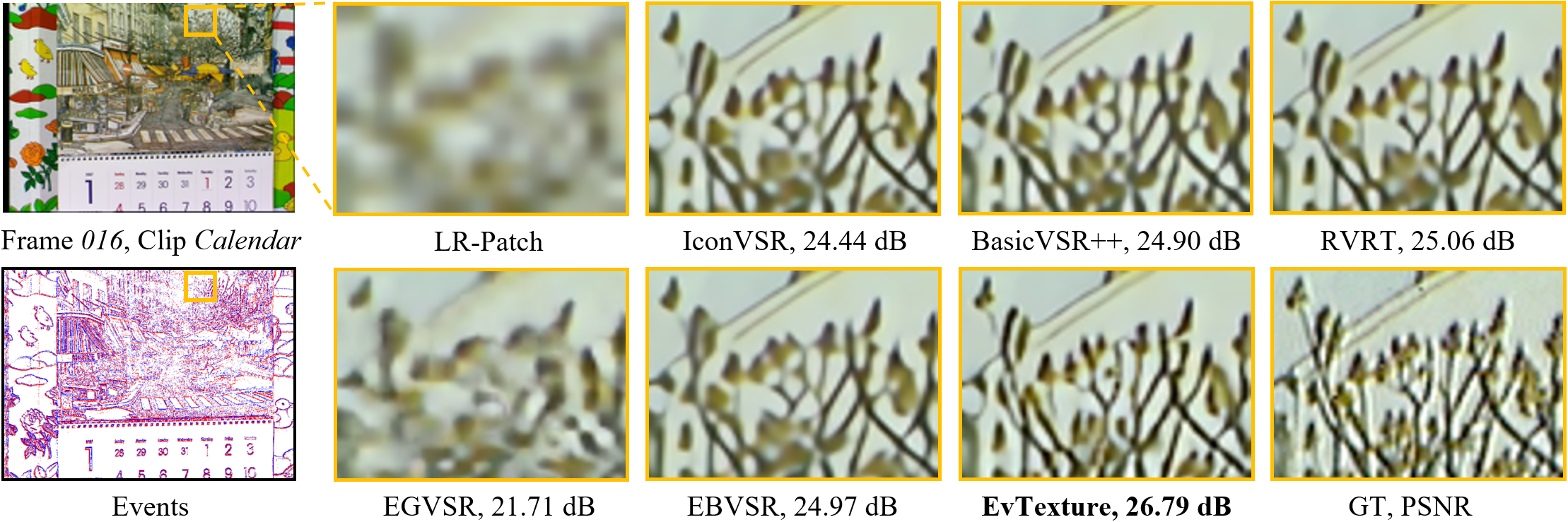}
        \caption{Qualitative comparison on Vid4~\cite{liu2013bayesian}. Our method can restore more vivid branches and leaves on the tulip tree.}
        \label{fig:fig4}
    \end{figure*}
    
    \begin{figure*}[t!]
        \centering
        \includegraphics[width=0.9\textwidth]{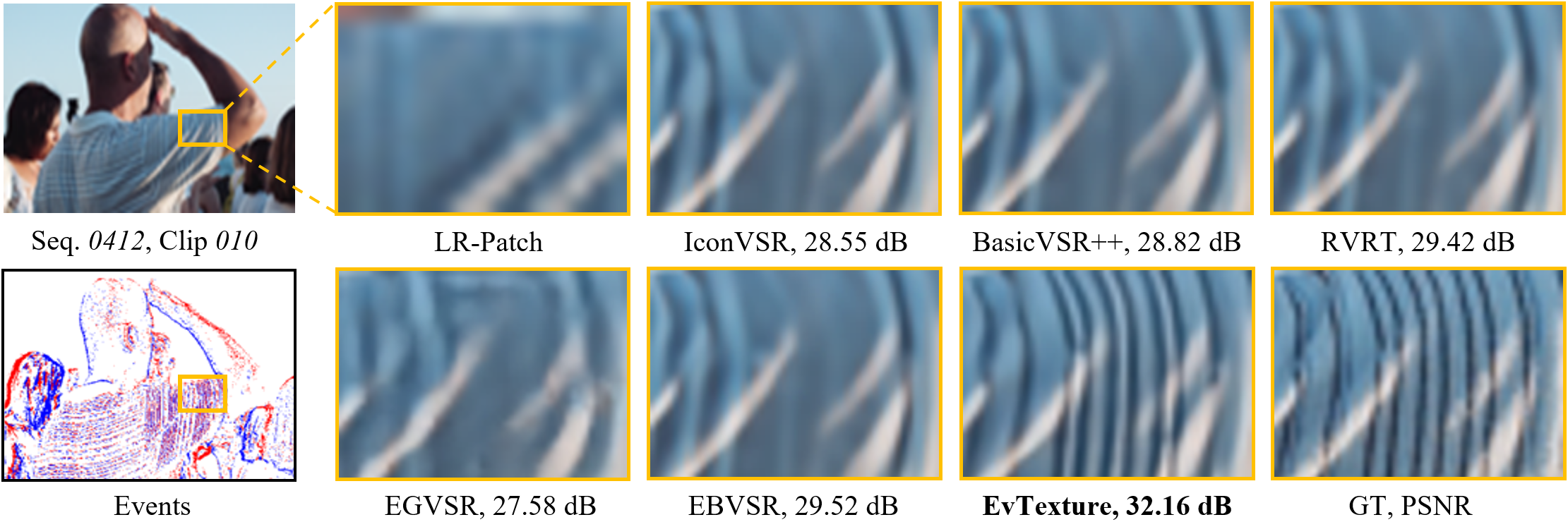}
        \caption{Qualitative comparison on Vimeo-90K-T~\cite{xue2019video}. Our method can restore more detailed stripes on clothing surfaces.}
        \label{fig:fig5}
    \end{figure*}

    \section{Experiments}
    \label{sec:experiments}
	
    \subsection{Datasets}
    \label{sec:datasets}
	
    \paragraph{Synthetic datasets.} We use two popular datasets for training: Vimeo-90K~\cite{xue2019video} and REDS~\cite{nah2019ntire}. Specifically, for Vimeo-90K, Vid4~\cite{liu2013bayesian} and Vimeo-90K-T serve as our test sets, assessing in the Y channel. For REDS, we employ REDS4\footnote{Clips 000, 011, 015, 020 of REDS training set.} as our test set and evaluate in the RGB channel. Notably, the Vimeo-90K, Vid4, and REDS datasets lack real event data. We follow the previous studies~\cite{jing2021turning,kai2023video}, and use the ESIM~\cite{rebecq2018esim} simulator to synthesize the corresponding events from clips. The simulated events are then converted to the voxel grid following Eqs.~(\ref{eq1}) and~(\ref{eq2}). Voxels are downsampled through bicubic interpolation, aligning with the frame downsampling method.

    \paragraph{Real-world datasets.} Following previous event-based VSR studies~\cite{lu2023learning,kai2023video}, we use the CED~\cite{scheerlinck2019ced} dataset for training and evaluating on real-world scenes. The dataset is captured with a DAVIS346~\cite{brandli2014real} event camera, which outputs temporally synchronized events and frames at a resolution of $260\times346$. Following~\cite{jing2021turning}, we select 11 clips\footnote{Scenes vary from static to dynamic, and indoor to outdoor.} from the total of 84 clips as our test set and use the remainder for training. When calculating the metrics, we exclude boundary 8 pixels and evaluate in the RGB channel.
	
    \subsection{Implementation Details}
    We use 15 frames as inputs for training and set the mini-batch size as 8 and the input frame size as $64\times64$. We augment the training data with random horizontal and vertical flips. We train the model for 300K iterations and adopt Adam optimizer~\cite{KingBa15} and Cosine Annealing~\cite{loshchilov2016sgdr} scheduler. The Charbonnier penalty loss~\cite{lai2017deep} is applied for supervision. A pre-trained model SpyNet~\cite{ranjan2017optical} is used to estimate optical flow, with other modules trained from scratch. For SpyNet, the initial learning rate is $2.5 \times 10^{-5}$, frozen for the first 5K iterations. The initial learning rate for other modules is $2 \times 10^{-4}$. The whole training is conducted on 8 NVIDIA RTX3090 GPUs and takes about four days.
	
    \subsection{Quantitative Results}
    We employ two types of baseline methods: RGB-based and event-based. For fair comparisons, we ensure that all methods are trained on the same dataset and evaluated under identical conditions. The evaluation metrics include widely-used PSNR and SSIM. Moreover, considering the common perception-distortion tradeoff~\cite{blau2018perception} problem in restoration algorithms, we additionally evaluate the perceptual similarity metric LPIPS~\cite{zhang2018unreasonable}, which aligns more closely with human perceptual cognition. We present the comparison results with other SOTA methods on four datasets in Tabs.~\ref{table:table1},~\ref{table:table2}, and~\ref{table:table3}. These results lead to several important conclusions.

    First, our EvTexture utilizes event signals more effectively than other event-based VSR methods. It achieves an impressive performance gain of \textbf{+4.67dB} in PSNR over the recent event-based method EGVSR on Vid4. Additionally, our method shows remarkable performance on the real-world CED dataset, achieving +1.83dB gain over EGVSR and +3.20dB improvement over E-VSR. From a perceptual quality perspective, the results in Tab.~\ref{table:table2}, Figs.~\ref{fig:fig4} and~\ref{fig:fig5} reveal that EvTexture can also offer superior perceptual similarity and restore more vivid and detailed texture regions.
	
    Second, event signals can provide high-frequency information that enhances RGB-based VSR. As shown in Tabs.~\ref{table:table1},~\ref{table:table2}, and~\ref{table:table3}, EvTexture attains significant improvements over RGB-based methods. For instance, compared to the recent best RGB-based model VRT, EvTexture requires fewer parameters and less runtime and achieves a +1.63dB increase on Vid4. Furthermore, our method surpasses the baseline model BasicVSR by +1.37dB on REDS4. Similarly, on CED, EvTexture outperforms BasicVSR by +0.95dB, further verifying the efficacy of incorporating event signals into the VSR framework. 
	
    \subsection{Qualitative Results}
    We also perform qualitative comparisons on these datasets. The visual comparisons are shown in Figs.~\ref{fig:fig4} and \ref{fig:fig5}. It is obvious that previous methods, whether using events or not, can not well restore the texture regions, leading to blurry textures or jitter effects. In contrast, our method excels in restoring detailed textures, such as tree branches and stripes on clothing surfaces, resulting in high-quality reconstructions. Sec.~\ref{sec:visuals} in the appendix provides more visual results, demonstrating the ability of our method to restore high-frequency textural details.

    \begin{figure}[t!]
        \centering
        \vspace{0.1cm}
        \includegraphics[width=0.95\columnwidth]{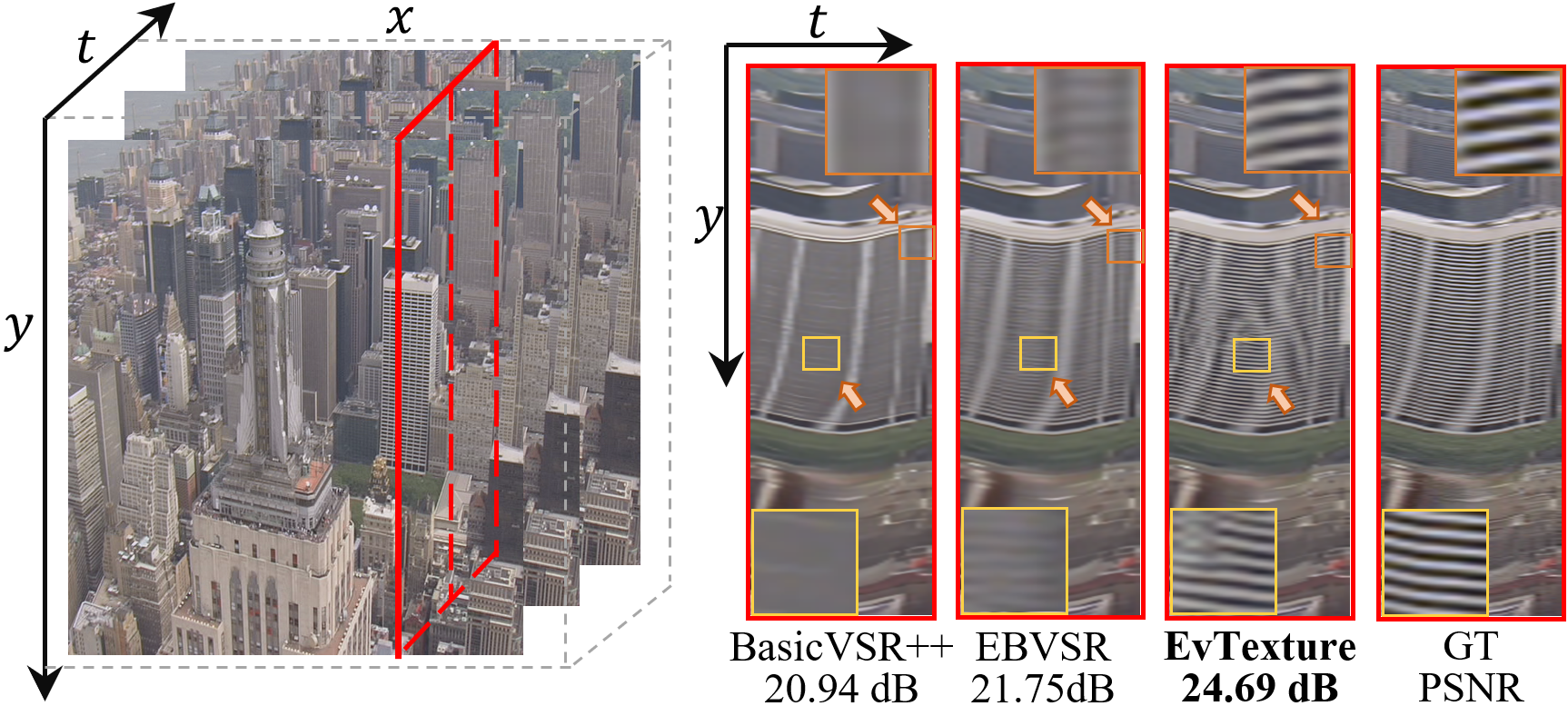}
        \caption{Comparison of temporal profile. We select a column (red dotted lines) to observe the changes across time. Our approach creates clearer and more consistent textures over time.}
        \label{fig:fig6}
    \end{figure}

        \begin{table}[t!]
    \vspace*{-0.1cm}
    \caption{Ablation studies of the two-branch structure. The texture branch brings significant improvements on Vid4 and REDS4.}
    \label{table:table4}
    \vskip 0.15in
    \centering
    \resizebox*{\columnwidth}{!}{
        \begin{tabular}{cccccc}
            \toprule 
            \multirow{2}[2]{*}{\makecell{Model \\ ID}} & \multicolumn{2}{c}{Branch} & \multirow{2}[2]{*}{\makecell{\#Params \\ (M)}}  & \multirow{2}[2]{*}{Vid4} &  \multirow{2}[2]{*}{REDS4} \\
            \cmidrule(lr){2-3} 
             & Motion & Texture &  &   &  \\ 
            \midrule
            (a) & \Checkmark & \XSolidBrush & 6.3 & 27.44/0.8284 & \textcolor{blue}{\underline{31.58}}/0.8932 \\
            (b) & \XSolidBrush & \Checkmark & 7.5  & \textcolor{blue}{\underline{29.22}}/\textcolor{blue}{\underline{0.8814}}  & 31.49/\textcolor{blue}{\underline{0.8942}} \\
            \textbf{EvTexture} & \Checkmark & \Checkmark & 8.9 & \textcolor{red}{\textbf{29.51}}/\textcolor{red}{\textbf{0.8909}} & \textcolor{red}{\textbf{32.79}}/\textcolor{red}{\textbf{0.9174}} \\
            \bottomrule
        \end{tabular}
    }
    \vspace*{-0.3cm}
    \end{table}
	
        \begin{table}[t!]
    \caption{Ablation studies about important factors of the Iterative Texture Enhancement module on Vid4.}
    \label{table:table5}
    \vskip 0.15in
    \centering
    \resizebox*{\columnwidth}{!}{
        \begin{tabular}{ccccccc}
            \toprule
            \multirow{2}[1]{*}{\makecell{Model \\ ID}} & \multirow{2}[1]{*}{\makecell{Texture \\ Updater}}  & \multirow{2}[1]{*}{\makecell{Iterative \\ Manner}} & \multirow{2}[1]{*}{\makecell{Residual \\ Learning}} & \multirow{2}[1]{*}{\makecell{Iteration \\ Number}} & \multirow{2}[1]{*}{\makecell{\#Params \\ (M)}} & \multirow{2}[1]{*}{\makecell{PSNR}}\\ [0.2em]
             &  &  & & &  &  \\
            \midrule
            (c) & Conv & \Checkmark & \Checkmark & 5 & 8.8 & 29.174 \\
            (d) & N/A & \XSolidBrush & \Checkmark & N/A & 7.6 & 29.087 \\
            (e) & ConvGRU & \Checkmark & \XSolidBrush & 5 & 8.5 & 29.158 \\
            \midrule
            (f) & ConvGRU & \Checkmark & \Checkmark & 3 & 8.9 & 29.463 \\
            (g) & ConvGRU & \Checkmark & \Checkmark & 8 & 8.9 & 29.324 \\
            \textbf{EvTexture} & ConvGRU & \Checkmark & \Checkmark &5& 8.9 & \textcolor{red}{\textbf{29.507}} \\
            \bottomrule
        \end{tabular}
    }
    \vskip -0.1in
    \end{table}

    \paragraph{Temporal Consistency.} We assess the temporal consistency of our inference results in texture regions using the temporal profile~\cite{xiao2020space,chan2022basicvsr++}. This method visualizes the temporal transition of frames by stacking selected pixel rows from each frame over time. As shown in Fig.~\ref{fig:fig6}, our method exhibits superior consistency in texture regions, attributed to our exploration of high-temporal-resolution events. Quantitatively, we achieve a +3.75dB gain over BasicVSR++ and a +2.94dB increase over the recent event-based model EBVSR. Our method not only restores clear textures spatially but also ensures smooth transitions temporally, closely similar to the ground truth.

    \begin{figure}[t!]
        \vspace{0.2cm}
        \centering
        \includegraphics[width=0.98\columnwidth]{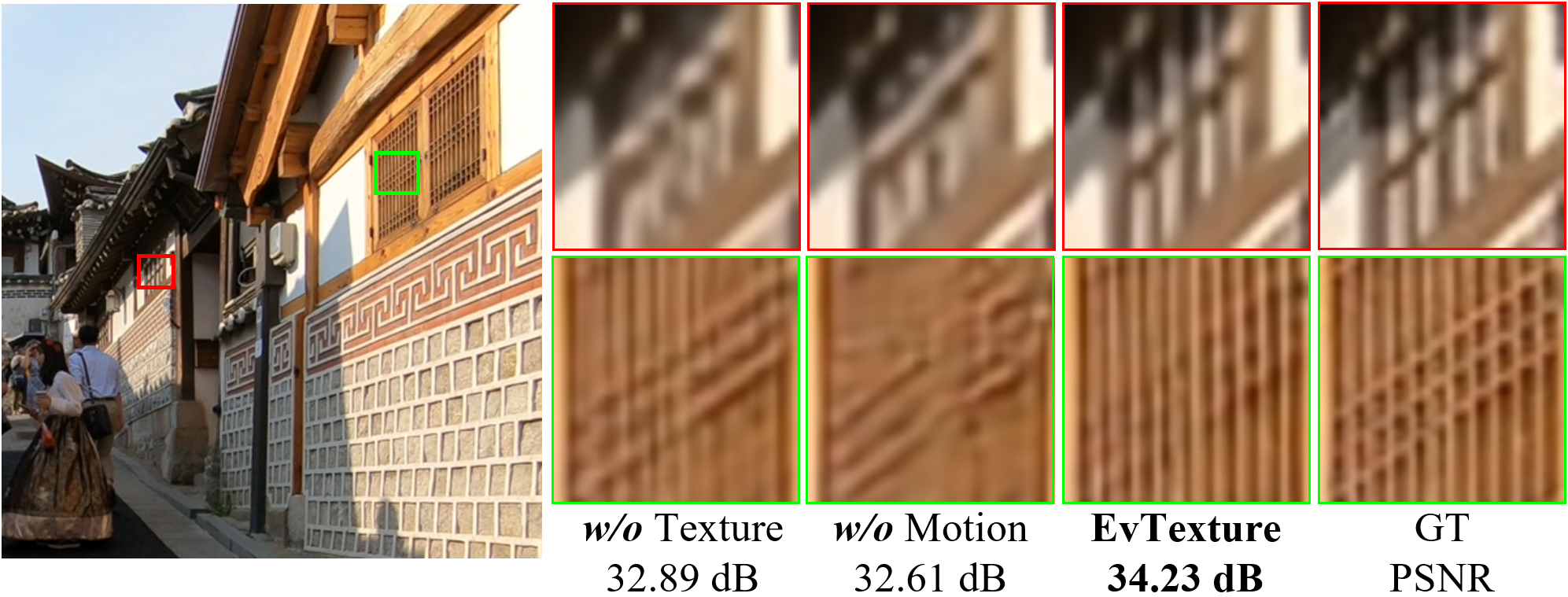}
        \caption{Ablations about the two-branch structure on REDS4. Our EvTexture can recover clearer grids on window surfaces. }
        \label{fig:fig7}
    \end{figure}
	
     \begin{figure}[t!]
        \centering
        \vspace{0.1cm}
        \includegraphics[width=0.98\columnwidth]{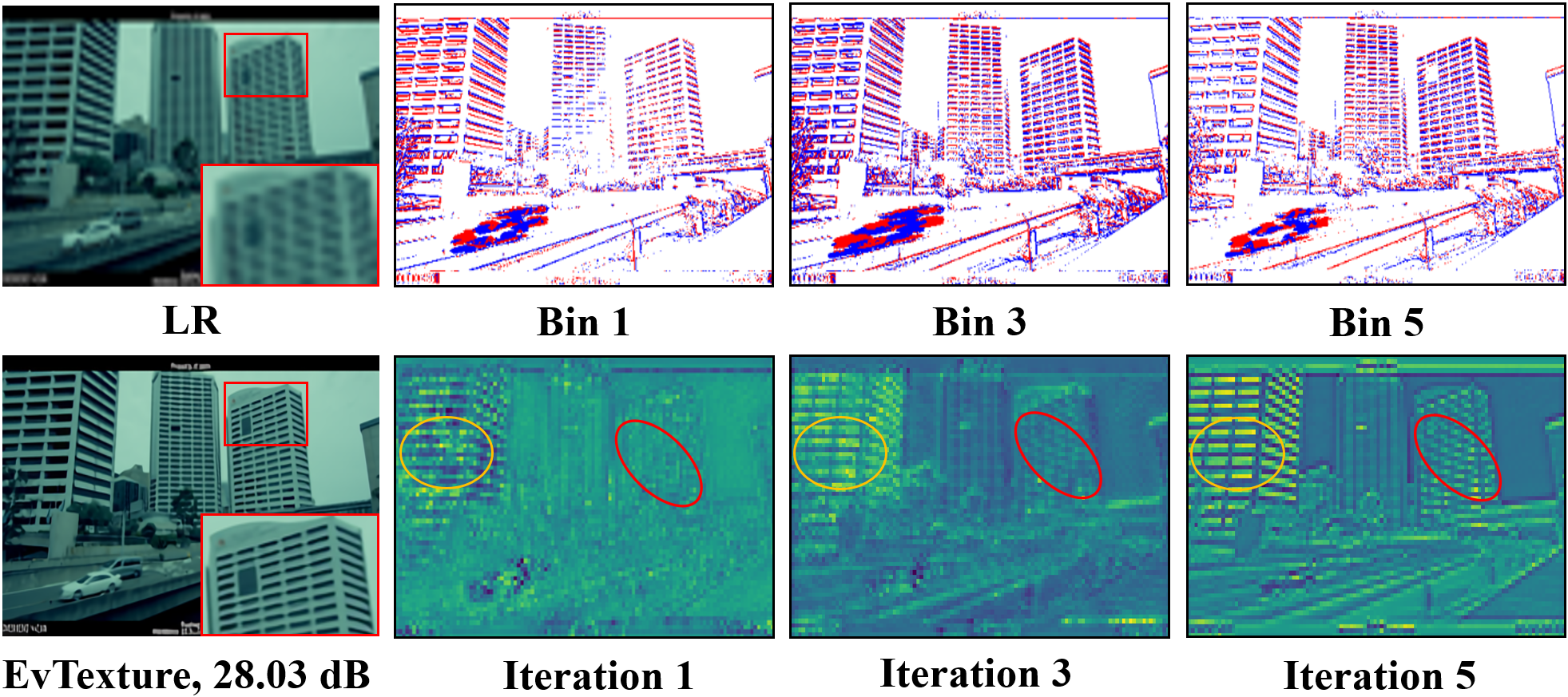}
        \caption{Analysis of iterative texture enhancement. As the iterations advance, the intermediate feature learns clearer texture details from voxel bins, progressively enhancing the restoration quality.}
        \label{fig:fig8}
        \vspace{-0.1cm}
    \end{figure}
	
    \subsection{Ablation Study}
    \label{subsec:abla}
       
    \paragraph{Two-branch Structure.} We first conduct ablations of each branch on the Vid4 and REDS4 datasets. The results are shown in Tab.~\ref{table:table4}. Here, Model (a) only equips the motion learning branch, and Model (b) features the texture enhancement branch. It suggests that on Vid4, Model (b) outperforms Model (a) by a margin of +2.07dB and achieves comparable performance to the EvTexture, indicating the dominant role of the texture branch. Moreover, on REDS4, our proposed model combining both branches achieves the best performance. Fig.~\ref{fig:fig7} shows visual comparisons, where our full model can restore more detailed window grids.
 
    \paragraph{Iterative Texture Enhancement.} We also examine some key factors in the ITE module in Tab.~\ref{table:table5}. Here, in Model (c), we replace the texture updater with three Conv layers. Our EvTexture with the ConvGRU block shows superior performance, as the gated activation better selects useful information for texture updating. In Model (d), instead of using an iterative update manner, we employ a specialized UNet to directly extract texture features from the whole voxel grid, which leads to a 0.42dB PSNR drop. For Model (e), we remove the residual learning approach used in Eqs.~(\ref{eq6}) and~(\ref{eq7}), which causes a 0.35dB PSNR drop. Models (f-g) and our EvTexture have different iteration numbers, and the results indicate that the ITE module with 5-iteration can achieve superior performance. More iterations are not necessary and may lead to worse performance, as there is significant high-frequency information loss in each bin.

    Additionally, we analyze the progression of the intermediate feature during iterations in Fig.~\ref{fig:fig8}. This analysis further demonstrates that as the texture updater advances, our EvTexture can adaptively learn to extract finer textures with less noise and more clarity from voxel bins for restoration.

     \begin{figure}[t!]
        \centering
        \includegraphics[width=\linewidth]{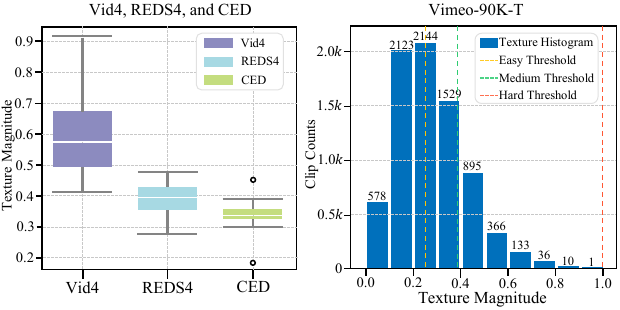}
        \vskip -0.1cm
        \caption{Texture magnitude analysis of four datasets. The Vid4 dataset has the most significant texture. The Vimeo-90K-T dataset exhibits a wide range of texture magnitudes, and we divide it into three difficulty levels (easy, medium, and hard).}
        \label{fig:fig9}
        \vspace{-0.3cm}
    \end{figure}

        \begin{table}[t!]
    \caption{Quantitative comparison (PSNR$\uparrow$/SSIM$\uparrow$) across easy, medium, and hard difficulty levels of Vimeo-90K-T for 4$\times$ VSR.}
    \label{table:table6}
    \vskip 0.15in
    \centering
    \resizebox{\columnwidth}{!}{
        \begin{tabular}{ l|ccc}
            \toprule
            \multirow{2}[1]{*}{Method} & \multicolumn{3}{c}{Vimeo-90K-T} \\ [-0.1em]
            \cmidrule(lr){2-4} 
            & Easy & Medium & Hard\\
            \midrule
            EDVR~\cite{wang2019edvr} & 41.98/0.975& 35.10/0.942 & 30.40/0.894\\
            BasicVSR~\cite{chan2021basicvsr}  & 41.55/0.973 & 34.63/0.937 & 29.97/0.886\\
            IconVSR~\cite{chan2021basicvsr} & 41.73/0.974 & 34.86/0.939 & 30.19/0.890\\
            BasicVSR++~\cite{chan2022basicvsr++} & 41.98/0.975 & 35.09/0.941 & 30.38/0.893 \\
            RVRT~\cite{liang2022recurrent} & 42.43/\textcolor{blue}{\underline{0.976}} & 35.69/0.946 & 30.73/\textcolor{blue}{\underline{0.902}} \\
            VRT~\cite{liang2022vrt} & \textcolor{red}{\textbf{42.47}}/\textcolor{blue}{\underline{0.976}} & 35.73/\textcolor{blue}{\underline{0.947}} & \textcolor{blue}{\underline{30.74}}/\textcolor{blue}{\underline{0.902}} \\
            EGVSR~\cite{lu2023learning} & 38.75/0.959 & 32.15/0.905 & 27.90/0.836 \\
            EBVSR~\cite{kai2023video} & 41.55/0.973 & 35.09/0.941 & 30.49/0.896 \\
            \textbf{EvTexture} & \textcolor{blue}{\underline{42.45}}/\textcolor{red}{\textbf{0.977}} & \textcolor{red}{\textbf{35.83}}/\textcolor{red}{\textbf{0.948}} & \textcolor{red}{\textbf{31.21}}/\textcolor{red}{\textbf{0.908}} \\
            \midrule
            $\#$ of clips & 3,907 & 2,345 & 1,563\\
            Avg. Texture Mag. & 0.16 & 0.32 & 0.49 \\
            \bottomrule
        \end{tabular}
    }
    \vspace{-0.3cm}
    \end{table}
	
    \section{Discussion}
    \subsection{Effectiveness of Texture Restoration}
    To further investigate the effectiveness of our method in texture restoration, and inspired by previous image texture analysis study~\cite{cai2022tdpn}, given a video that has $T$ frames of size $H\times W$, we calculate the texture magnitude of a video clip as follows:
    
    \begin{equation}\label{eq:eq9}
        \frac{\alpha}{T}\sum_{t=1}^{T}\sqrt{\frac{1}{HW}\sum_{i=1}^{H}\sum_{j=1}^{W}\left|I_t(i,j) - \bar{I}_t(i,j)\right|^2}.
    \end{equation}
 
    Here, we first smooth each frame with a Gaussian filter to obtain the blurred image $\bar{I}$, with a kernel size of $(5, 5)$ and $\sigma=1.5$. Then, we calculate the absolute difference between the original and blurred images to extract high-frequency textural details, and compute the average contrast across the sequence. $\alpha$ is a scaling factor, and in our experiments, we set $\alpha=10$. The texture magnitude in Eq.~(\ref{eq:eq9}) ranges from 0 to 1, where a higher value indicates the clip contains richer textures and is more difficult to restore. 

    Accordingly, we analyze the texture magnitude of the four datasets mentioned in Sec.~\ref{sec:datasets}, with the results presented in Fig.~\ref{fig:fig9}. It reveals that the Vid4 dataset has the most significant texture, followed by REDS4, and then CED. It is worth noting that Vimeo-90K-T has a large amount of 7,815 clips. We categorize these clips into three difficulty levels: easy, medium, and hard, based on each clip's texture magnitude and our careful empirical observations. We then compare our method with SOTA methods across these three difficulty level subsets in Tab.~\ref{table:table6}. 

    The results suggest a meaningful conclusion that our EvTexture especially excels in texture-rich datasets and clips. For instance, on two datasets with different texture levels, Vid4 and CED, Tabs.~\ref{table:table1} and~\ref{table:table3} reveal that EvTexture surpasses the recent event-based method, EBVSR, by +1.05dB and +0.38dB, respectively. Moreover, Tab.~\ref{table:table6} shows that our method achieves comparable performance to VRT on the easy subset of Vimeo-90K-T, while our method is more effective on the hard subset, achieving a gain of up to +0.47dB.
	
    \subsection{Extension to EvTexture+} Furthermore, despite our primary focus on texture restoration in VSR, we also draw insights from previous event-based flow estimation studies~\cite{wan2022learning,shiba2022secrets}. They exploit the motion information from events to enhance the optical flow, so as to boost the inter-frame alignment like~\cite{kai2023video}. Our EvTexture can easily adapt to these approaches. Thus, we extend it to \textbf{EvTexture+}, further utilizing events to enhance the motion learning in VSR. Its architecture is detailed in Fig.~\ref{fig:figA3} in the appendix. The results in Tabs.~\ref{table:table1},~\ref{table:table2} and~\ref{table:table3} show that EvTexture+ can bring further performance gains over EvTexture. In some cases, such as on the Vid4 dataset, we notice that the gains from EvTexture+ are minor, which indicates that our EvTexture is already highly effective in texture restoration. 
	
    \section{Conclusion}
    This paper presents EvTexture, a novel event-driven texture enhancement network dedicated to texture restoration in VSR by incorporating high-frequency event signals. Our model is based on a recurrent architecture, and we propose a two-branch structure in which a parallel texture enhancement branch is introduced in addition to the motion branch. Furthermore, we propose an iterative texture enhancement module to progressively enhance the texture details through multiple iterations. Experimental results show that our EvTexture outperforms existing SOTA methods and especially excels in restoring finer textures. In the future, we will focus on 1) evaluating our method under more challenging scenarios, including fast-moving and low-light conditions, and 2) extending our approach to handle asymmetrical spatial resolutions between frames and events, which are more common in practical applications.

    \section*{Acknowledgements}
	
    This work was in part supported by the National Natural Science Foundation of China under Grant 62021001.  

    \section*{Impact Statement} 

    This paper presents work whose goal is to advance the field of Machine Learning. There are many potential societal consequences of our work, none which we feel must be specifically highlighted here.

    % \clearpage \newpage
    \bibliography{example_paper}
    \bibliographystyle{icml2024}

	%%%%%%%%%%%%%%%%%%%%%%%%%%%%%%%%%%%%%%%%%%%%%%%%%%%%%%%%%%%%%%%%%%%%%%%%%%%%%%%
	%%%%%%%%%%%%%%%%%%%%%%%%%%%%%%%%%%%%%%%%%%%%%%%%%%%%%%%%%%%%%%%%%%%%%%%%%%%%%%%
	% APPENDIX
	%%%%%%%%%%%%%%%%%%%%%%%%%%%%%%%%%%%%%%%%%%%%%%%%%%%%%%%%%%%%%%%%%%%%%%%%%%%%%%%
	%%%%%%%%%%%%%%%%%%%%%%%%%%%%%%%%%%%%%%%%%%%%%%%%%%%%%%%%%%%%%%%%%%%%%%%%%%%%%%%
	%%%%%%%%%%%%%%%%%%%%%%%%%%%%%%%%%%%%%%%%%%%%%%%%%%%%%%%%%%%%%%%%%%%%%%%%%%%%%%%
%%%%%%%%%%%%%%%%%%%%%%%%%%%%%%%%%%%%%%%%%%%%%%%%%%%%%%%%%%%%%%%%%%%%%%%%%%%%%%%
% APPENDIX
%%%%%%%%%%%%%%%%%%%%%%%%%%%%%%%%%%%%%%%%%%%%%%%%%%%%%%%%%%%%%%%%%%%%%%%%%%%%%%%
%%%%%%%%%%%%%%%%%%%%%%%%%%%%%%%%%%%%%%%%%%%%%%%%%%%%%%%%%%%%%%%%%%%%%%%%%%%%%%%

\newpage
\appendix
\onecolumn
\definecolor{Customblue}{RGB}{0,20,115}
\textcolor{white}{dasdsa}
\section*{\textcolor{Customblue}{\Large{Appendix Contents}}}
\setcounter{tocdepth}{2}  
% Customize the table of contents for sections
\titlecontents{section}[0em]{\color{Customblue}\bfseries}{\thecontentslabel. }{}{\hfill\contentspage}

% Customize the table of contents for subsections
\titlecontents{subsection}[1.5em]{\color{Customblue}}{\thecontentslabel. }{}{\titlerule*[0.75em]{.}\contentspage} % Adjust the space as needed

% Customization for Appendix contents
\renewcommand{\contentsname}{Appendix Contents}

\begingroup
\let\clearpage\relax
\startcontents[sections]
\printcontents[sections]{l}{1}{}
\newpage
\endgroup

\newpage
\section{Texture Restoration Challenge}
\label{section:problem analysis}

\begin{figure}[H]
	\vspace*{-0.2cm}
	\centering
	\subfigure[\textbf{Significant errors concentrate in the woven basket area.}] {
		\includegraphics[width=0.86\textwidth]{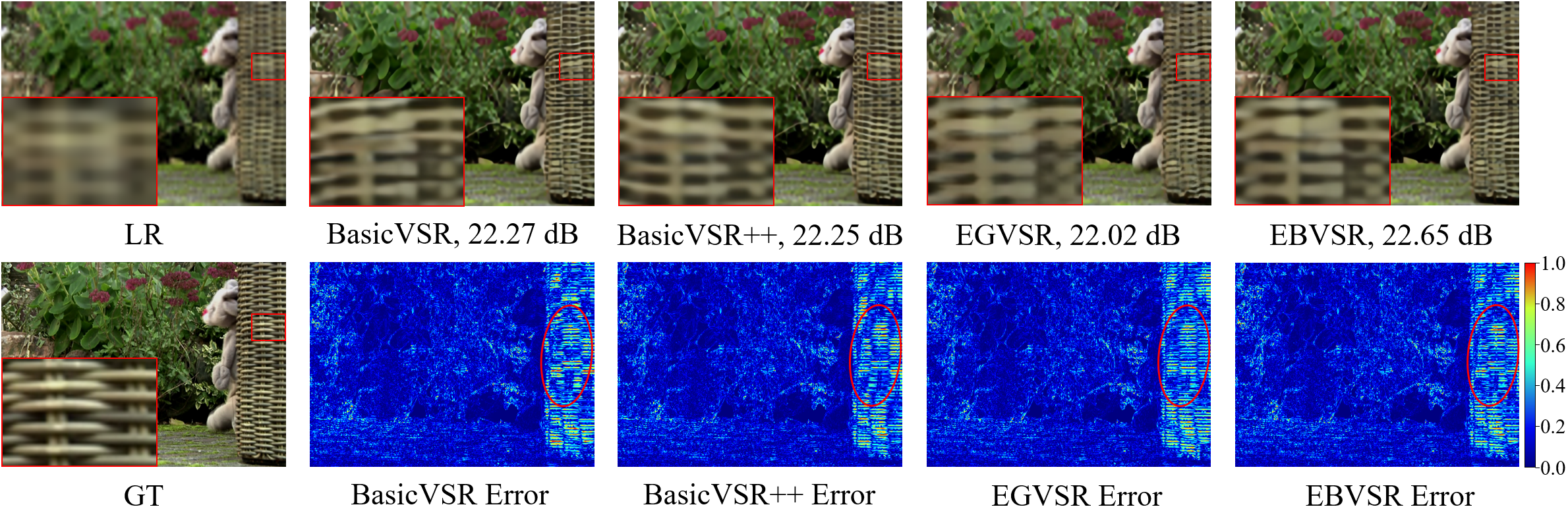}
	}
	\subfigure[\textbf{Large errors focus on building surfaces.}] {
		\includegraphics[width=0.86\textwidth]{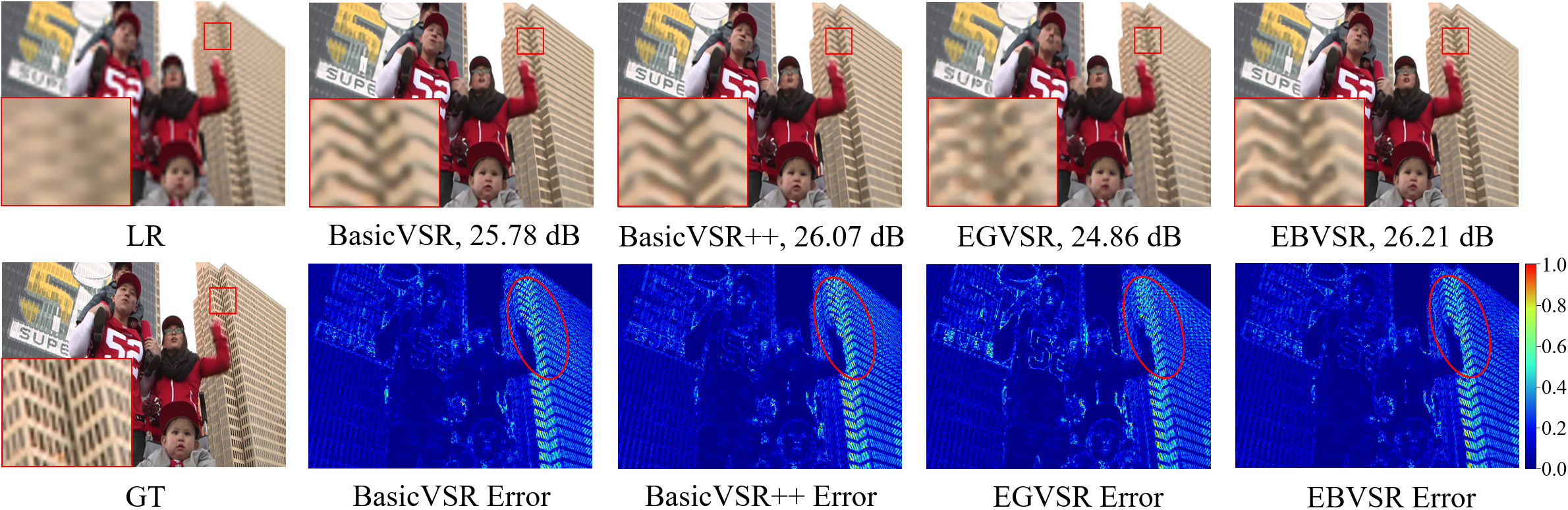}
	}
	\subfigure[\textbf{Considerable errors occur in texture-rich areas, such as tree branches and leaves.}] {
		\includegraphics[width=0.86\textwidth]{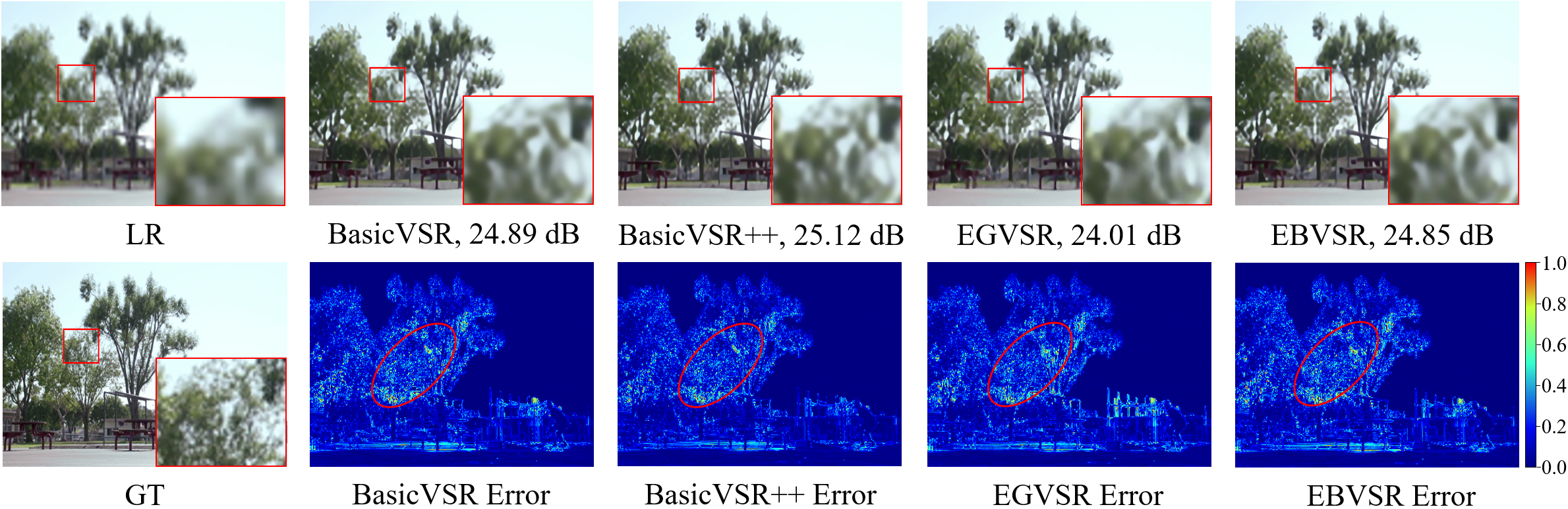}
	}
	\caption{\textbf{Problem analysis.} Existing state-of-the-art VSR methods, such as RGB-based BasicVSR~\cite{chan2021basicvsr} and BasicVSR++~\cite{chan2022basicvsr++}, and event-based EGVSR~\cite{lu2023learning} and EBVSR~\cite{kai2023video}, still suffer from blurry textures, leading to noticeable errors in texture regions. \textbf{Zoomed in for best view.}}
	\label{fig:figA1}
\end{figure}

\vspace{-0.1cm}
Texture restoration is a very challenging problem in VSR. It is hard to predict rich texture details in HR videos from the corresponding LR ones. It is also difficult to preserve the temporal consistency of the texture regions during video playback, given the rich details. As shown in Fig.~\ref{fig:figA1}, we provide more examples to illustrate the texture restoration challenge in VSR. We analyze two representative RGB-based methods, BasicVSR~\cite{chan2021basicvsr} and BasicVSR++~\cite{chan2022basicvsr++}, as well as two recent event-based methods, EGVSR~\cite{lu2023learning} and EBVSR~\cite{kai2023video}. These methods produce blurry textures and exhibit large errors in texture-rich areas such as fabrics, building surfaces, and natural landscapes.

\clearpage

\section{Motivation about Our Work}
\label{section:motivation}

\begin{figure}[H]
	\vspace*{-2ex}
	\centering
	\subfigure[Event generation and representation] {
		\includegraphics[width=0.58\textwidth]{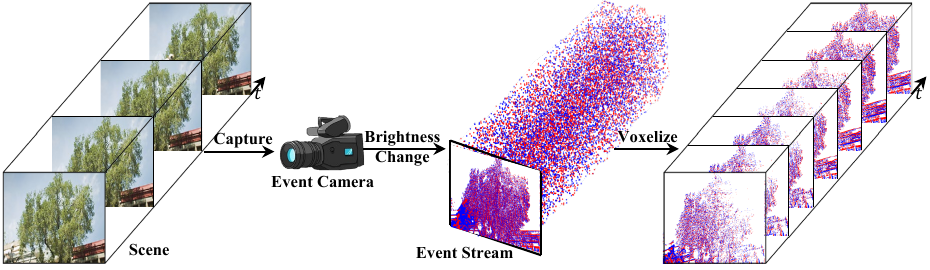}
	}
	\subfigure[Comparison of event-based VSR] {
		\includegraphics[width=0.37\textwidth]{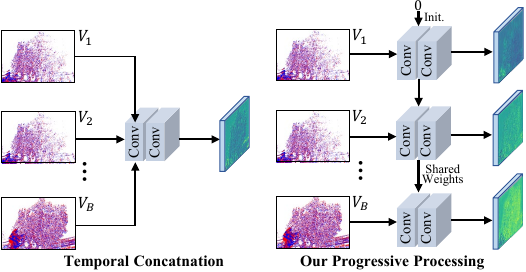}
	}
	\vspace*{-2ex}
	\caption{Analysis of event signals and our method. (a) Event cameras asynchronously capture scene brightness changes and output a high-temporal-resolution event stream. We represent these events as voxel grids proposed by Zhu~\etal~\yrcite{zhu2019unsupervised}. (b) Previous event-based VSR methods~\cite{jing2021turning, lu2023learning, kai2023video} directly concatenate voxel bins temporally. In contrast, our method processes bins progressively, enhancing textures iteratively and preserving the temporal relationships among these bins.}
	\label{fig:figA2}
	\vspace*{-2ex}
\end{figure}

\subsection{Event Generation Model}
Event cameras respond to changes in the logarithmic brightness signal $L\left(u_k, t_k\right) \doteq \log I\left(u_k, t_k\right)$ asynchronously and independently~\cite{gallego2020event}. An event is triggered at pixel $u_k=\left(x_k,y_k\right)$ and at time $t_k$ as soon as the brightness changes over a pre-setting contrast threshold $\pm C$ ($C>0$) as:
\begin{equation}
	L\left(u_k, t_k\right)-L\left(u_k, t_k-\Delta t\right) \geq p_k C,
\end{equation}
where $p_k \in\{+1,-1\}$ is the polarity of brightness change (increase or decrease), and $\Delta t$ is the time since the last event at $u_k$. Consequently, an event is made up of $(x_k,y_k,t_k,p_k)$. The output event stream from an event camera can be denoted as a set of 4-tuples $\mathcal{E}\in \mathbb{R}^{N_e \times 4}$, where $N_e$ represents the number of events.

\subsection{High-frequency Textural Information in Events} Fig.~\ref{fig:figA2}(a) depicts the event generation and representation model. Events can indeed provide rich high-frequency textural information to enhance VSR results. The reasons are three-fold. 

\begin{itemize}
	\item First, event cameras measure per-pixel brightness changes, and such changes usually occur first at the edges of objects due to object movements. This type of ``moving edge" information has been referenced in event-based segmentation tasks~\cite{mitrokhin2020learning} and is extensively utilized in event-assisted deblurring studies~\cite{sun2022event}.
	\item Second, unlike standard cameras, which rely on a fixed clock, event cameras asynchronously sample light based on scene dynamics~\cite{gallego2020event}. This enables rapid and repeated responses to movements, allowing for multiple samplings in a short period. Thus, events have richer edge information compared to standard frame-based signals.
	\item Third, it is well-known that edges represent high-frequency information, and rich edges can assist in recognizing textural patterns. As a result, events can provide high-frequency textural information for enhancing VSR.
\end{itemize}

\subsection{Progressive Processing of Events} Fig.~\ref{fig:figA2}(b) depicts our approach in utilizing high-temporal-resolution events. Unlike methods that directly concatenate voxel bins over time, we iteratively enhance texture regions by breaking optimization into steps: $t,\ t+\delta,\ t+2\delta,\ ...\ ,t+1$, where $\delta=1/B$. This is achieved through our iterative texture updater, which utilizes ConvGRU units. The benefits of this approach are three-fold:
\begin{itemize}
	\item The iterative structure allows for progressive enhancement of textural details across multiple iterations, resulting in a more precise restoration of the texture regions.
	\item Instead of directly passing the event data to synthesize texture regions all at once, our structure also models the temporal relationships within events.
	\item ConvGRU units can utilize information from previous steps (hidden state) to influence current decisions.
\end{itemize}

\clearpage

\section{Extension to EvTexture+}
\label{section:EvTexture+}

Texture restoration is a very challenging problem in VSR, and our EvTexture is designed to focus on solving this problem with the help of high-frequency textural information from events. Additionally, we also draw insights from previous event-based flow estimation studies~\cite{wan2022learning,shiba2022secrets} to exploit motion information of events to enhance the flow estimation and temporal alignment. We thus develop the \textbf{EvTexture+}, an extension of our EvTexture, further utilizing events to enhance the motion learning in VSR. 

\begin{figure}[H]
	\centering
	\includegraphics[width=0.95\textwidth]{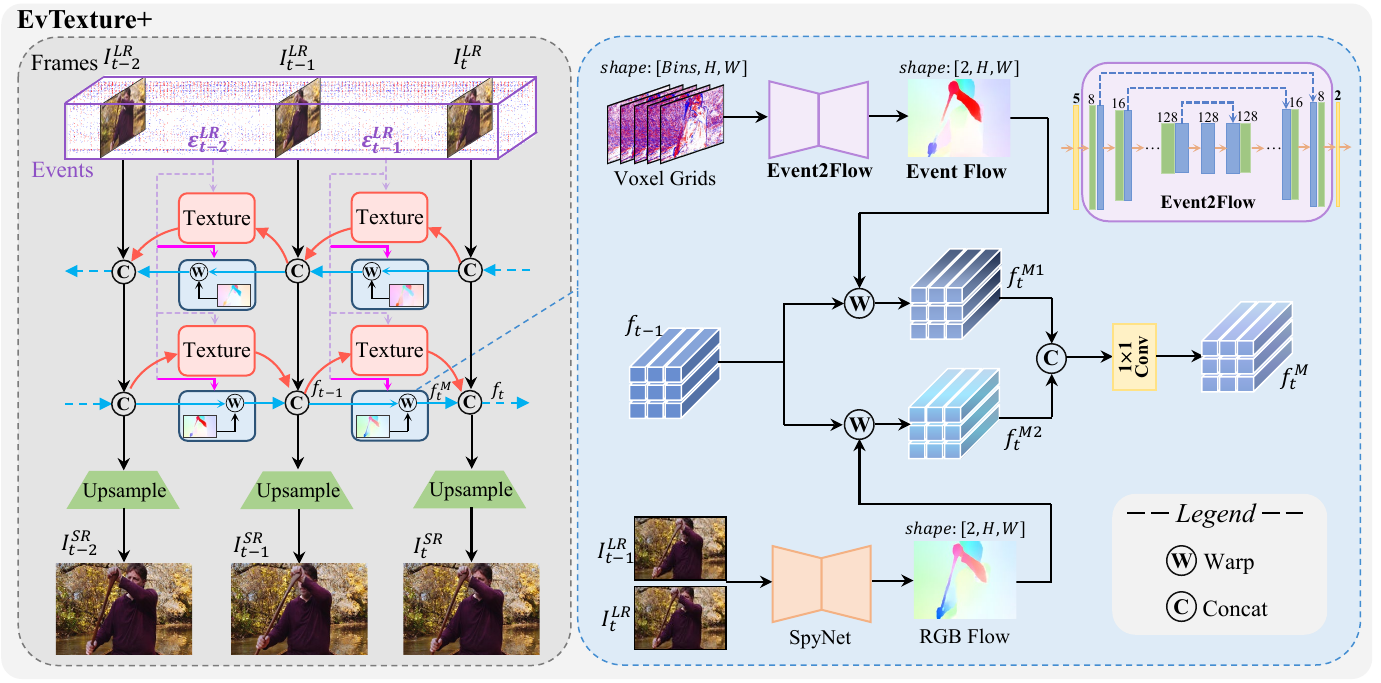}
	\caption{Network architecture of EvTexture+. Compared to EvTexture, it additionally incorporates events to enhance motion learning.}
	\label{fig:figA3}
\end{figure}

Fig.\ref{fig:figA3} presents the network architecture of \textbf{EvTexture+}. The left part of the figure shows that it includes an additional data flow (\textbf{\textcolor{magenta}{magenta}} arrow) from events to the motion branch. In the motion branch, events are first converted to the voxel grid as described in Eqs~(\ref{eq1}) and~(\ref{eq2}). We design an Event2Flow network to capture nonlinear motion properties in events. This network, similar to EBVSR~\cite{kai2023video}, is a custom U-Net~\cite{ronneberger2015u} and generates event-driven optical flow. Afterwards, event flow and RGB flow separately warp $f_{t-1}$ to obtain $f_t^{M1}$ and $f_t^{M2}$, respectively. $f_t^{M1}$ incorporates rich nonlinear motion information from events, while $f_t^{M2}$ includes color and content information from frames. These two features are fused by a $1\times1$ Conv to create the motion-enhanced feature $f_t^M$. It reveals that EvTexture+ can readily adapt to existing event-based VSR methods, which primarily exploit the motion information from events.

The results presented in Tabs.~\ref{table:table1},~\ref{table:table2}, and ~\ref{table:table3} suggest that EvTexture+ achieves additional performance improvements over EvTexture in most cases. In certain cases, for example, on the Vid4 dataset in Tab.~\ref{table:table1}, the gains from EvTexture+ are minor, which suggests that our EvTexture is already highly effective in texture restoration.

\clearpage

\section{Full Experimental Results}
\label{sec:fullResults}

In this section, we present detailed clip-by-clip test results for the REDS4~\cite{nah2019ntire} and CED~\cite{scheerlinck2019ced} datasets, shown in Tabs.~\ref{table:table_a1} and~\ref{table:table_a2}. Note that the last column of each table presents the clip’s texture magnitude, calculated via Eq.~(\ref{eq:eq9}). A higher Texture Magnitude (Mag.) value indicates that the clip has richer textures and is more difficult to restore.

    \begin{table}[H]
    \caption{Clip-by-clip results (PSNR$\uparrow$/SSIM$\uparrow$) on REDS4~\cite{nah2019ntire} for 4$\times$ VSR. Results are evaluated on the RGB-channel. \textcolor{red}{\textbf{Red}} indicates the best performance. \textcolor{blue}{\underline{Underlined}} indicates some significant values that reflect our method's superiority with rich texture clips.}
    \label{table:table_a1}
    \vskip 0.15in
    \centering
    \resizebox{\textwidth}{!}{
    \begin{tabular}{cccccccccc} 
    \toprule
    \multirow{3}[3]{*}{\makecell{\textbf{REDS4}\\[0.2ex]Clip Name}} & \multicolumn{3}{c}{RGB-based VSR} & \multicolumn{3}{c}{Event-based VSR} & \multirow{3}[3]{*}{\makecell{ EvTexture\\\textit{vs.} VRT}} & \multirow{3}[3]{*}{\makecell{ EvTexture\\\textit{vs.} EBVSR}} & \multirow{3}[2]{*}{\makecell{Texture\\ Mag. \\ (Eq.~(\ref{eq:eq9}))}} \\
        \cmidrule(lr){2-4} \cmidrule(lr){5-7} 
        & \multirow{2}[1]{*}{\makecell{BasicVSR\\ \small{\cite{chan2021basicvsr}}}} & \multirow{2}[1]{*}{\makecell{TTVSR\\ \small{\cite{liu2022learning}}}} & \multirow{2}[1]{*}{\makecell{VRT\\ \small{\cite{liang2022vrt}}}} & \multirow{2}[1]{*}{\makecell{EGVSR\\ \small{\cite{lu2023learning}}}} & \multirow{2}[1]{*}{\makecell{EBVSR\\ \small{\cite{kai2023video}}}}  & \multirow{2}[1]{*}{\textbf{EvTexture}} & & \\[3ex]
    
        \midrule
        000 & 28.40/0.8434 & 28.82/0.8565 & 28.85/0.8553 & 25.16/0.7066 & 28.44/0.8446 & \textcolor{red}{\textbf{30.72}}/\textcolor{red}{\textbf{0.9082}} & \textcolor{blue}{\underline{+1.87/+0.0529}} & \textcolor{blue}{\underline{+2.28/+0.0636}} & \textcolor{blue}{\underline{0.47}} \\
    
        011 & 32.47/0.8979 & 33.46/0.9100 & 33.49/0.9072 & 26.56/0.7722 & 32.55/0.8987 & \textcolor{red}{\textbf{33.72}}/\textcolor{red}{\textbf{0.9145}} & +0.23/+0.0073 & +1.17/+0.0158 & 0.38 \\
    
        015 & 34.18/0.9224 & 35.01/0.9325 & \textcolor{red}{\textbf{35.26}}/\textcolor{red}{\textbf{0.9332}} & 29.83/0.8526 & 34.22/0.9235 & 35.06/0.9314 & \textcolor{blue}{\underline{-0.20/-0.0018}} & \textcolor{blue}{\underline{+0.84/+0.0079}} & \textcolor{blue}{\underline{0.29}} \\
    
        020 & 30.63/0.9000 & 31.17/0.9093 & 31.16/0.9078 & 25.94/0.7846 & 30.67/0.9009 & \textcolor{red}{{\textbf{31.65}}}/\textcolor{red}{\textbf{0.9154}} & +0.49/+0.0076 &  +0.98/+0.0145 & 0.41 \\
    
        \midrule
        Average & 31.42/0.8909 & 32.12/0.9021 & 32.19/0.9006 & 26.87/0.7790 & 31.47/0.8919 & \textcolor{red}{\textbf{32.79}}/\textcolor{red}{\textbf{0.9174}} & +0.60/+0.0168 &  +1.32/+0.0255 & 0.39 \\
        \bottomrule
    \end{tabular}
    }
    \vskip -0.1in
    \end{table}
    \begin{table}[H]
    \caption{Clip-by-clip results (PSNR$\uparrow$/SSIM$\uparrow$) on CED~\cite{scheerlinck2019ced} for 2$\times$ VSR. Results are evaluated on the RGB-channel. $^\dagger$ denotes results are from EGVSR~\cite{lu2023learning}.}
    \label{table:table_a2}
    \vskip 0.15in
    \centering
    \resizebox*{\textwidth}{!}{
        \begin{tabular}{cccccccccc}
            \toprule
            
            \multirow{3}[3]{*}{\makecell{\textbf{CED}\\[0.2ex]Clip Name}} & \multicolumn{3}{c}{RGB-based VSR} & \multicolumn{3}{c}{Event-based VSR} & \multirow{3}[3]{*}{\makecell{ EvTexture\\\textit{vs.} BasicVSR}} & \multirow{3}[3]{*}{\makecell{ EvTexture\\\textit{vs.} EBVSR}} & \multirow{3}[2]{*}{\makecell{Texture\\ Mag. \\ (Eq.~(\ref{eq:eq9}))}} \\
            \cmidrule(lr){2-4} \cmidrule(lr){5-7} 
            
             & \multirow{2}[1]{*}{\makecell{TDAN$^\dagger$\\ \small{\cite{tian2020tdan}}}} & \multirow{2}[1]{*}{\makecell{RBPN$^\dagger$\\ \small{\cite{haris2019recurrent}}}} & \multirow{2}[1]{*}{\makecell{BasicVSR\\ \small{\cite{chan2021basicvsr}}}} & \multirow{2}[1]{*}{\makecell{EGVSR$^\dagger$\\ \small{\cite{lu2023learning}}}} & \multirow{2}[1]{*}{\makecell{EBVSR\\ \small{\cite{kai2023video}}}} & \multirow{2}[1]{*}{\textbf{EvTexture}} & & \\[3ex]
            \midrule 
            
            people\_dynamic\_wave  & 35.83/0.9540 & 40.07/\textcolor{red}{\textbf{0.9868}} & 39.35/0.9784  & 38.78/0.9794 & 39.95/0.9811 & \textcolor{red}{\textbf{40.39}}/0.9824 & +1.04/+0.0040 & +0.44/+0.0013 & 0.34\\
            
            indoors\_foosball\_2 & 32.12/0.9339 & 34.15/0.9739 & 39.81/0.9766 & 38.68/0.9750 & 40.23/0.9780 & \textcolor{red}{\textbf{40.54}}/\textcolor{red}{\textbf{0.9789}} & +0.73/+0.0023 & +0.31/+0.0009 & 0.30 \\
            
            simple\_wires\_2 & 31.57/0.9466 & 33.83/0.9739 & 39.73/0.9832 & 38.67/0.9815 & 40.31/0.9849 & \textcolor{red}{\textbf{40.75}}/\textcolor{red}{\textbf{0.9859}} & +1.02/+0.0027 & +0.44/+0.0010 & 0.34\\
            
            people\_dynamic\_dancing & 35.73/0.9566 & 39.56/\textcolor{red}{\textbf{0.9869}} & 39.60/0.9789  & 39.06/0.9798 & 40.22/0.9816  & \textcolor{red}{\textbf{40.66}}/0.9829 & +1.06/+0.0040 & +0.44/+0.0014 & 0.33 \\
            
            people\_dynamic\_jumping & 35.42/0.9536 & 39.44/\textcolor{red}{\textbf{0.9859}} & 39.45/0.9778 & 38.93/0.9792 & 40.06/0.9805  & \textcolor{red}{\textbf{40.45}}/0.9819 & +1.00/+0.0041 & +0.39/+0.0004 & 0.34\\
            
            simple\_fruit\_fast & 37.75/0.9440 & 40.33/0.9782 & 42.71/0.9815 & 41.96/0.9831 & 43.08/0.9830 & \textcolor{red}{\textbf{43.27}}/\textcolor{red}{\textbf{0.9834}} & \textcolor{blue}{\underline{+0.56/+0.0019}} & \textcolor{blue}{\underline{+0.19/+0.0004}} & \textcolor{blue}{\underline{0.18}}\\
            
            outdoor\_jumping\_infrared\_2 & 28.91/0.9062 & 30.36/0.9648 & 39.15/0.9748 & 38.03/0.9755 & 39.97/0.9754 & \textcolor{red}{\textbf{40.53}}/\textcolor{red}{\textbf{0.9800}} & \textcolor{blue}{\underline{+1.38/+0.0052}} & \textcolor{blue}{\underline{+0.56/+0.0046}} & \textcolor{blue}{\underline{0.44}}\\
            
            simple\_carpet\_fast & 32.54/0.9006 & 34.91/0.9502 & 36.97/0.9672& 36.14/0.9635 & 37.24/0.9689 & \textcolor{red}{\textbf{37.57}}/\textcolor{red}{\textbf{0.9705}} & +0.60/+0.0033 & +0.33/+0.0016 & 0.36 \\
            
            people\_dynamic\_armroll & 35.55/0.9541 & 40.05/\textcolor{red}{\textbf{0.9878}} & 39.35/0.9776 & 38.84/0.9787 & 39.95/0.9802 & \textcolor{red}{\textbf{40.35}}/0.9815 & +1.00/+0.0039 & +0.40/+0.0013 & 0.34 \\
    
            indoors\_kitchen\_2 & 30.67/0.9323 & 31.51/0.9551 & 38.45/0.9732 & 37.68/0.9726 & 38.88/0.9757  & \textcolor{red}{\textbf{39.27}}/\textcolor{red}{\textbf{0.9769}} & +0.82/+0.0037 & +0.39/+0.0012 & 0.33 \\
    
            people\_dynamic\_sitting & 35.09/0.9561 & 39.03/\textcolor{red}{\textbf{0.9862}} & 39.41/0.9799 & 38.86/0.9810 & 40.07/0.9804 & \textcolor{red}{\textbf{40.54}}/0.9837 & +1.13/+0.0038 & +0.47/+0.0033 & 0.36 \\
    
            \midrule 
            Average & 33.74/0.9398 & 36.66/0.9754 & 39.57/0.9778 & 38.69/0.9771 & 40.14/0.9801  & \textcolor{red}{\textbf{40.52}}/\textcolor{red}{\textbf{0.9813}} & +0.95/+0.0035 & +0.38/+0.0012 & 0.33\\
            
            \bottomrule
        \end{tabular}
    }
    \vskip -0.1in
    \end{table}

\subsection{Clip-by-clip Results on REDS4}

Tab.~\ref{table:table_a1} highlights that our EvTexture performs best on clips `000', `011', and `020' on REDS4. In particular, on the most texture-rich clip, `000', EvTexture outperforms VRT and EBVSR, achieving PSNR improvements of +1.87dB and +2.28dB, respectively. Conversely, on the texture-weak clip `015', our method is 0.20dB lower than VRT and only +0.84dB higher than EBVSR. These results demonstrate our method's superiority, especially in scenes with rich textures.

\subsection{Clip-by-clip Results on CED}

Tab.~\ref{table:table_a2} suggests our EvTexture achieves the best performance in most scenarios. Notably, on the clip with the richest textures, `outdoor\_jumping\_infrared\_2', EvTexture surpasses EBVSR by +0.56dB, while on the weaker texture clip, `simple\_fruit\_fast', the gain is only +0.19dB. A similar situation also occurs when comparing EvTexture with BasicVSR. The gain of our method is greater on clips with rich textures and less on clips with weaker textures.

\clearpage

\section{Performance versus Runtime}
\label{sec:performance}

\begin{figure}[htbp]
	\centering
	\includegraphics[width=0.95\textwidth]{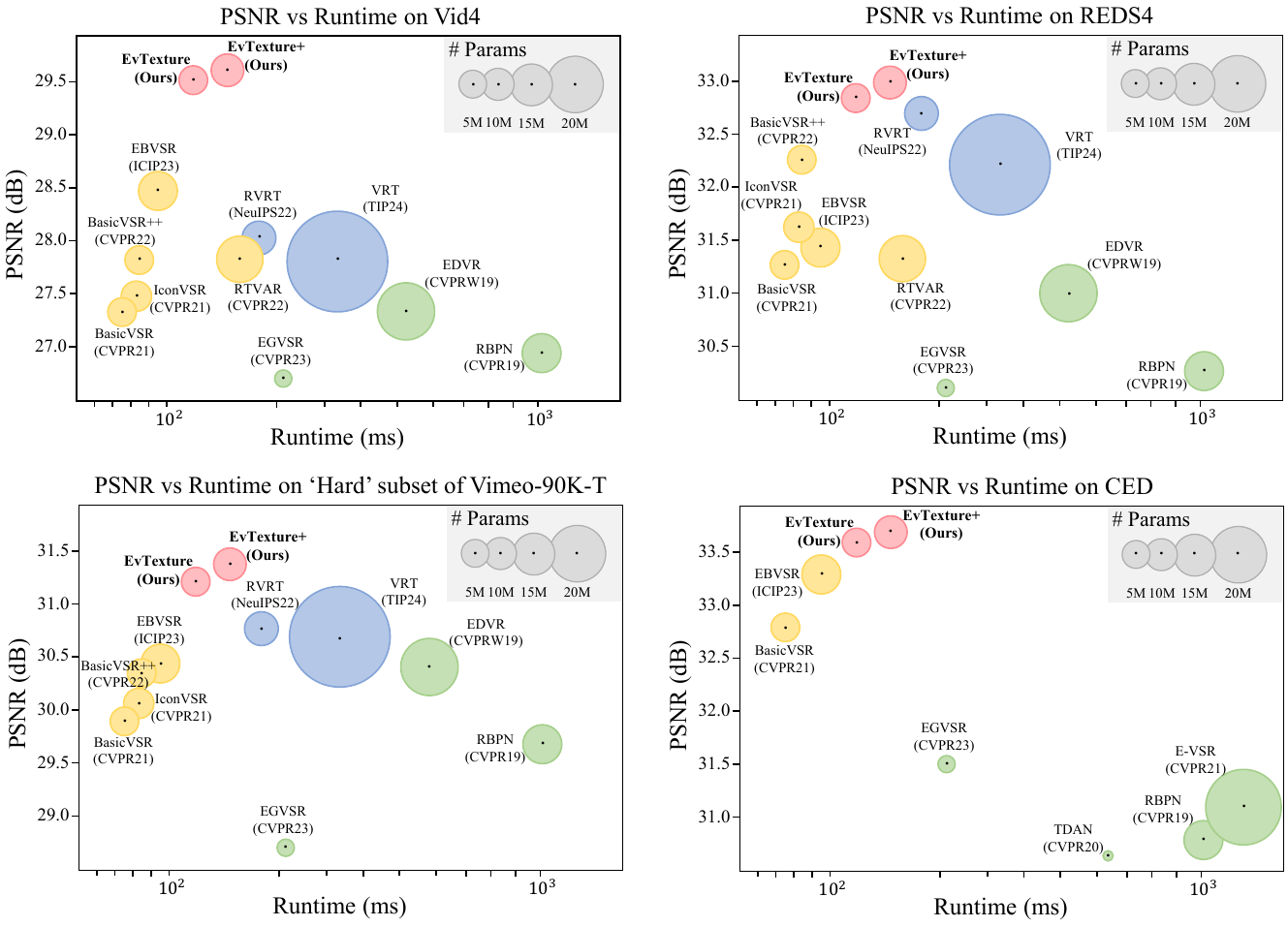}
	\caption{Performance gain on four datasets for $4\times$ VSR. The yellow circles represent RNN-based methods, while the blue ones indicate transformer-based methods. The green circles denote sliding-window approaches, which output only one frame at a time, resulting in longer runtime. Our method is based on an RNN-based VSR approach BasicVSR~\cite{chan2021basicvsr}.}
	\label{fig:figA11}
	\vskip -0.1in
\end{figure}

Fig.~\ref{fig:figA11} shows plots comparing performance (PSNR), runtime, and number of parameters on four test sets. Our EvTexture outperforms other state-of-the-art methods on these datasets, offering a better balance in terms of performance, parameters, and runtime. On the texture-rich dataset Vid4, our EvTexture and EvTexture+ outperform BasicVSR++ by +1.64dB and +1.91dB, respectively, with only a minimal cost increase in parameters and runtime.

\vspace{0.5cm}
\section{More Visual Results}
\label{sec:visuals}

\begin{figure}[htbp]
	\centering
	\includegraphics[width=0.9\textwidth]{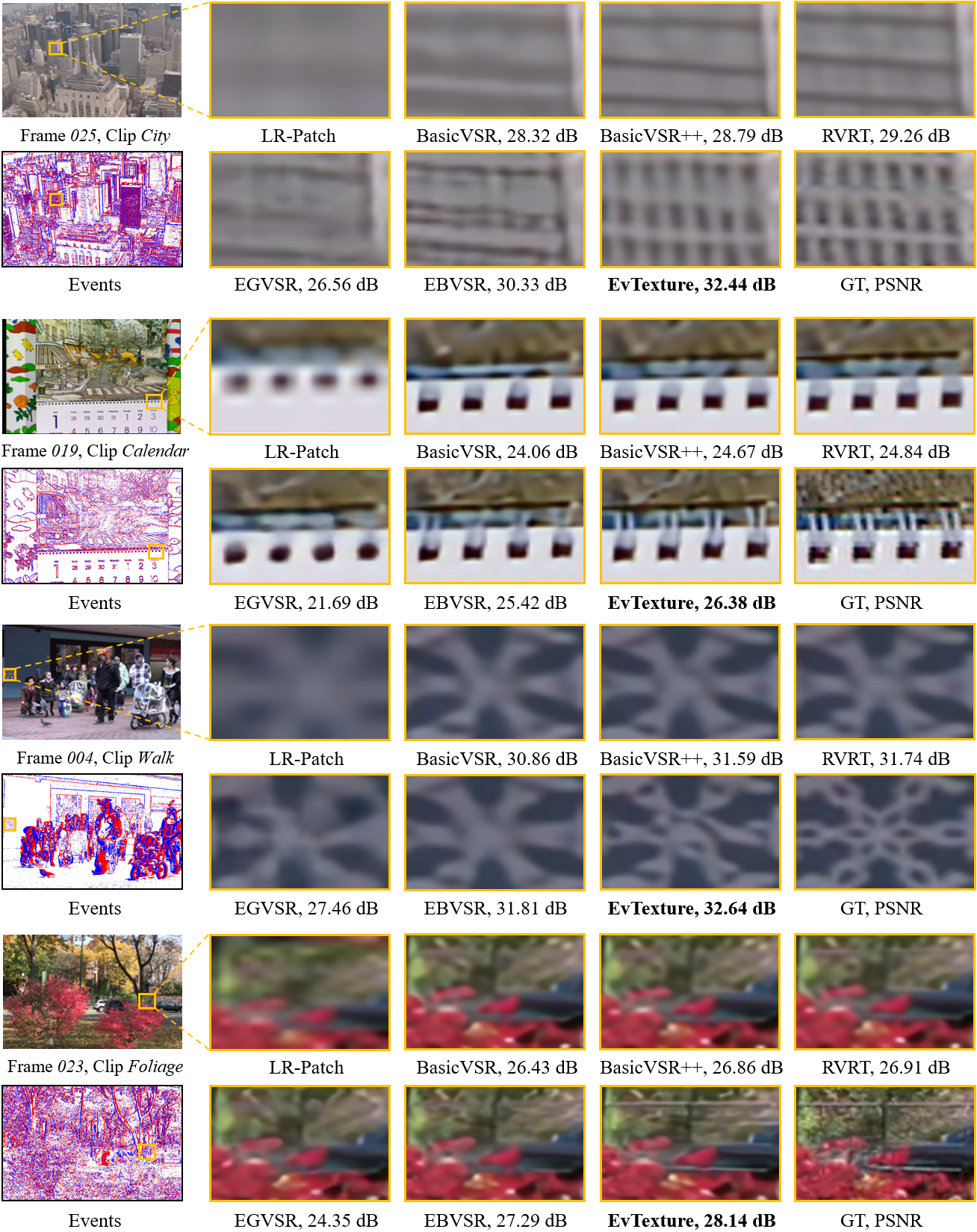}
	\caption{Qualitative comparison on Vid4~\cite{liu2013bayesian} for 4$\times$ VSR. \textbf{Zoomed in for best view.}}
	\label{fig:figA5}
	\vskip -0.2in
\end{figure}

\begin{figure}[htbp]
	\centering
	\includegraphics[width=0.9\textwidth]{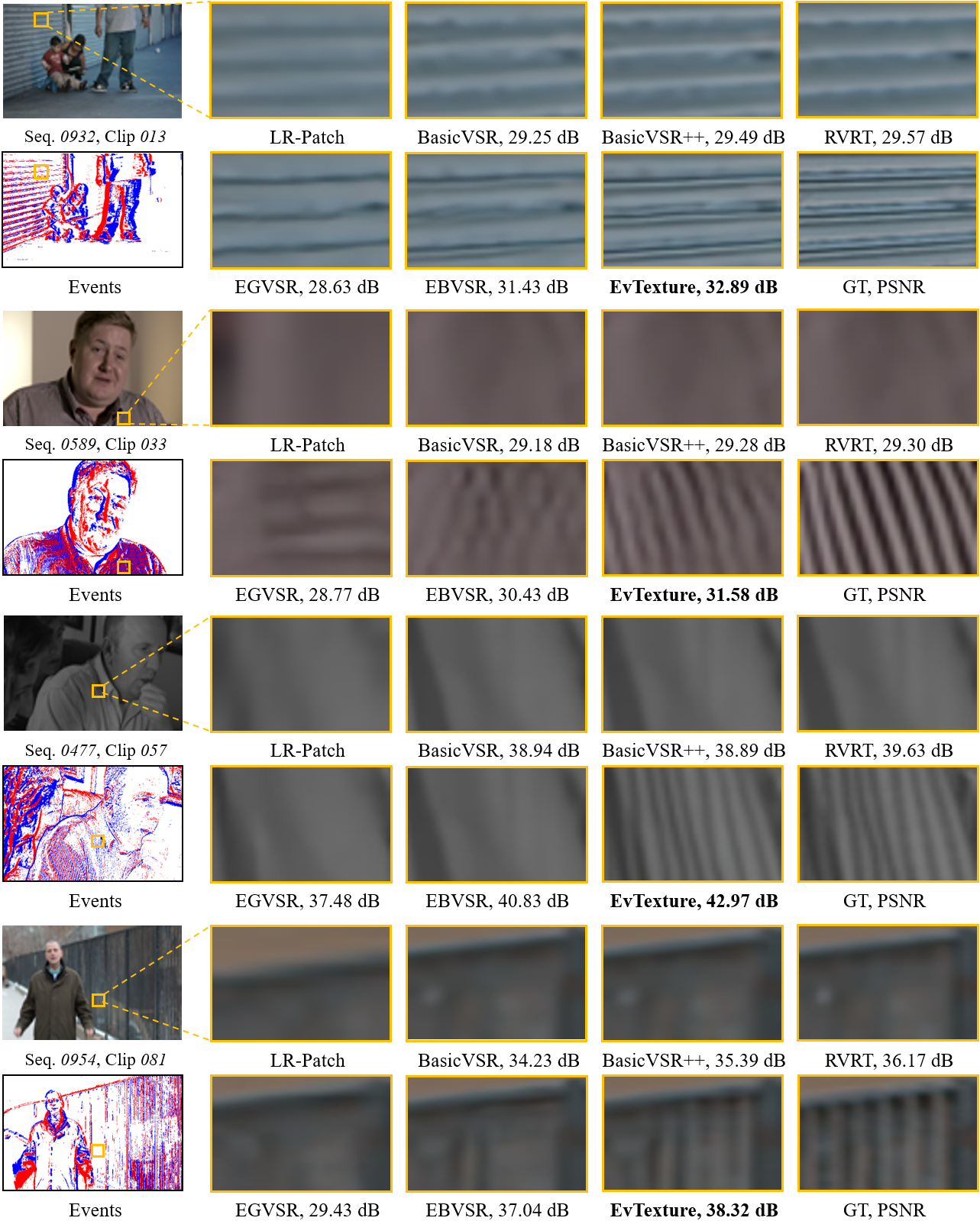}
	\caption{Qualitative comparison on Vimeo-90K-T~\cite{xue2019video} `Hard' for 4$\times$ VSR. \textbf{Zoomed in for best view.}}
	\label{fig:figA6}
	\vskip -0.2in
\end{figure}

\begin{figure}[htbp]
	\centering
	\includegraphics[width=0.9\textwidth]{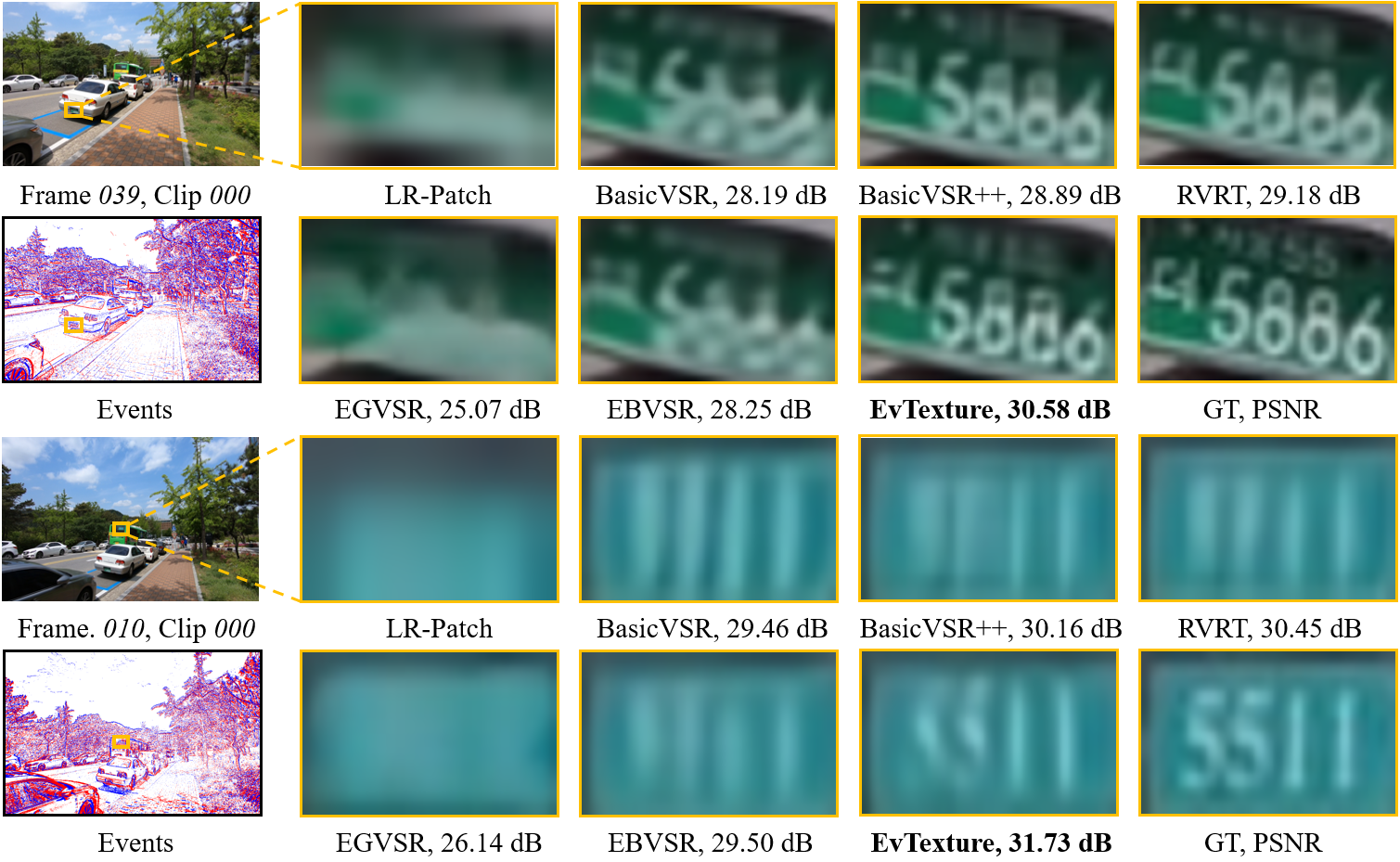}
	\caption{Qualitative comparison on REDS4~\cite{nah2019ntire} for 4$\times$ VSR. \textbf{Zoomed in for best view.}}
	\label{fig:figA7}
	\vskip -0.2in
\end{figure}

\begin{figure}[htbp]
	\centering
	\includegraphics[width=0.9\textwidth]{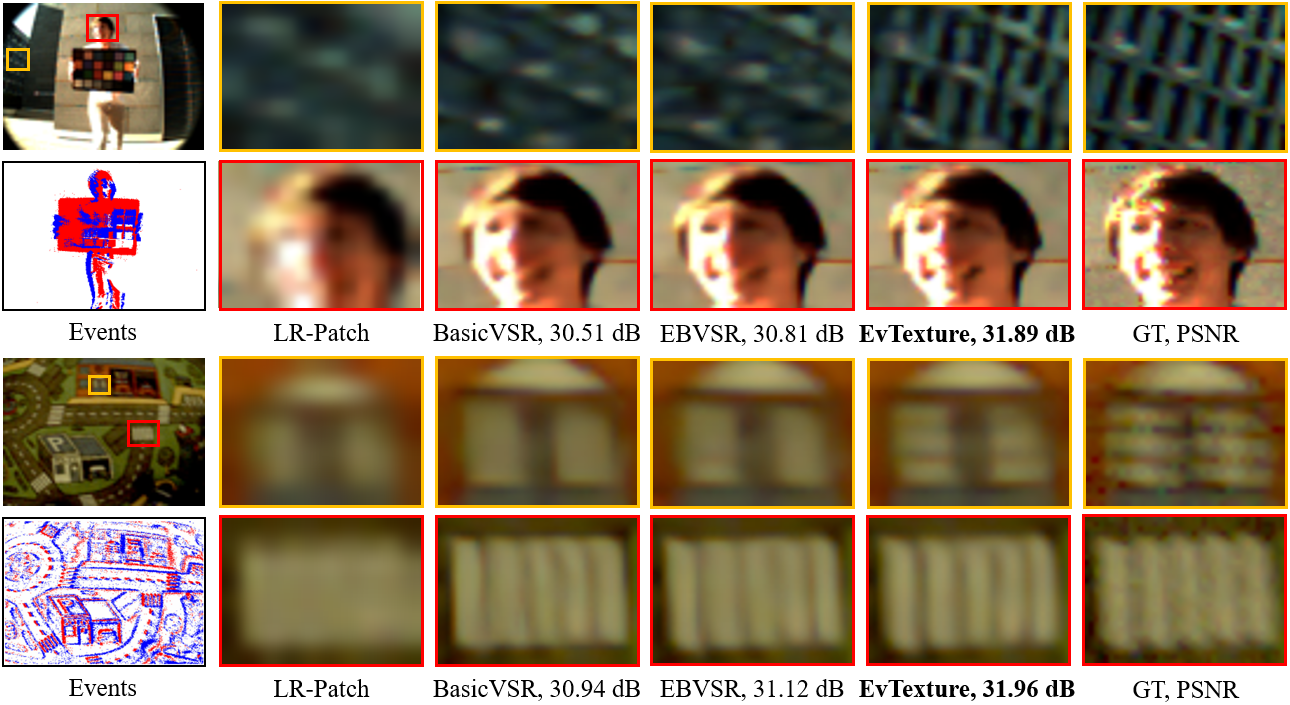}
	\caption{Qualitative comparison on CED~\cite{scheerlinck2019ced} for 4$\times$ VSR. \textbf{Zoomed in for best view.}}
	\label{fig:figA8}
	\vskip -0.2in
\end{figure}

In this section, to further validate the performance of our EvTexture in restoring texture regions, we provide additional visual comparisons on Vid4 \cite{liu2013bayesian}, Vimeo-90K-T \cite{xue2019video}, REDS4 \cite{nah2019ntire}, and CED \cite{scheerlinck2019ced} datasets. The results are shown in Figs.~\ref{fig:figA5},~\ref{fig:figA6},~\ref{fig:figA7} and~\ref{fig:figA8}, respectively. These results show that our EvTexture can successfully restore more vivid and detailed textures across various scenes, including the patterns on woven fabrics, clothes, building surfaces, and the textures of tree branches and leaves. 

\clearpage

\section{Dataset Details}
\label{sec:dataset}

\subsection{Synthetic Dataset}

\subsubsection{Vid4}
\label{sec:vid4}

The Vid4~\cite{liu2013bayesian} dataset is one of the most popular test datasets for VSR. It consists of four video sequences: `calendar' (41 frames with a resolution of $576\times720$), `city' (34 frames with a resolution of $576\times704$), `foliage' (49 frames with a resolution of $480\times720$), and `walk' (47 frames with a resolution of $480\times720$). This dataset provides diverse scenes, making it ideal for assessing VSR algorithms. The results on this dataset are shown in Tabs .~\ref{table:table1} and~\ref{table:table2}.

\subsubsection{REDS}

The REDS~\cite{nah2019ntire} dataset, proposed in the NTIRE 2019 Challenge, is a high-quality (720p) video dataset often used for video deblurring and super-resolution tasks. It consists of 270 video sequences, divided into 240 sequences for training and 30 for validation. Each sequence features 100 consecutive frames with a resolution of $720\times1280$. We follow EDVR~\cite{wang2019edvr} and select four representative clips, known as REDS4\footnote{Clips 000, 011, 015, 020 of REDS training set.}, which offers diverse scenes and motions as our test set. The remaining clips are regrouped to form our training dataset, comprising 266 clips. The evaluation results on this dataset are shown in Tabs.~\ref{table:table1},~\ref{table:table2} and~\ref{table:table_a1}.

\subsubsection{Vimeo-90K}

The Vimeo-90K~\cite{xue2019video} dataset is a large-scale, high-quality video dataset, with 91,701 sequences of 7 frames each at $256\times448$ resolution. These sequences are extracted from around 39,000 video clips, providing a comprehensive resource for video processing research. The dataset includes 64,612 training clips and 7,824 testing clips, known as Vimeo-90K-T. Following RBPN~\cite{haris2019recurrent}, we remove nine clips from Vimeo-90K-T due to their all-black backgrounds.

We then calculate the texture magnitude of the remaining 7,815 clips using Eq.~(\ref{eq:eq9}). As shown in Fig.~\ref{fig:fig9}, Vimeo-90K-T has a wide range of texture magnitudes. Thus, we categorize these clips into three levels: easy, medium, and hard. Specifically, we first sort the clips in ascending order based on their texture magnitudes. After comprehensive user studies and empirical observations, we classify the first 50\% (3,907 clips) as easy, the next 30\% (2,345 clips) as medium, and the final 20\% (1,563 clips) as hard. While our division percentages may not be the most precise, we hope this dataset can provide a better way to study and evaluate texture-related approaches. The test results on three difficulty level subsets are shown in Tab.~\ref{table:table6}.

\subsubsection{Simulating Events}

Since the Vid4, REDS, and Vimeo-90K datasets lack real event data, we follow the approach widely used in event-based frame interpolation~\cite{tulyakov2021time,kim2023event}, deblurring~\cite{sun2022event} and VSR~\cite{jing2021turning,kai2023video} studies. Accordingly, we use the ESIM~\cite{rebecq2018esim} simulator to generate synthetic event data. The ESIM simulator takes a video clip as input and generates corresponding event streams. Before simulating events, we employ the widely used video frame interpolation model RIFE~\cite{huang2022real} to create a high frame rate video with $8\times$ interpolation scale. When simulating events, we apply a threshold $c$ that follows a Gaussian distribution $N(\mu=1,\sigma=0.1)$ to accurately mimic the dynamics of real-world scenes.

\subsection{Real-world Dataset}

\subsubsection{CED}

Following event-based VSR studies~\cite{jing2021turning,lu2023learning,kai2023video}, we use CED~\cite{scheerlinck2019ced} as our real-world event dataset. It comprises a collection of color event streams and video sequences, totaling 84 clips in various scenes such as indoor, outdoor, driving, and human moving. We follow the preprocessing steps of E-VSR~\cite{jing2021turning} to use 73 clips for training and the remaining 11 clips for testing. The resolution of both frames and events is $260\times346$. It is important to note that some clips contain a very large amount of frames, for example, 10,541 frames in the `driving\_city\_5' clip. Considering memory limitations, processing these frames simultaneously is challenging. Therefore, we process every 15 frames during training, and for testing, we perform inference on every 100 frames. The results on this dataset are shown in Tabs.~\ref{table:table3} and~\ref{table:table_a2}.

\clearpage

\section{More Discussions}
\label{sec:discussion}

\textbf{Discussion 1: Why are Vid4 results in Tab.~\ref{table:table1} different from original papers?}

The results of previous methods in Tab.~\ref{table:table1} show slight differences from those reported in their original papers. For instance, the original paper of BasicVSR++~\cite{chan2022basicvsr++} reports an average result of 27.79 dB in terms of PSNR on the Vid4 dataset, whereas, in our Tab.~\ref{table:table1}, we report their result as 27.87 dB. This discrepancy arises because they calculated the average result by treating the four clips of Vid4 as having the same number of frames. However, as mentioned earlier in Sec.~\ref{sec:vid4}, \textbf{these four clips have different lengths}. Thus, we recalculated their average results, considering the varying clip lengths. 

\textbf{Discussion 2: Is the texture magnitude calculation in Eq.~(\ref{eq:eq9}) reasonable?}

\begin{figure}[H]
	\vspace*{-2ex}
	\centering
	\subfigure[Feasibility of texture magnitude calculation] {
		\includegraphics[width=0.55\textwidth]{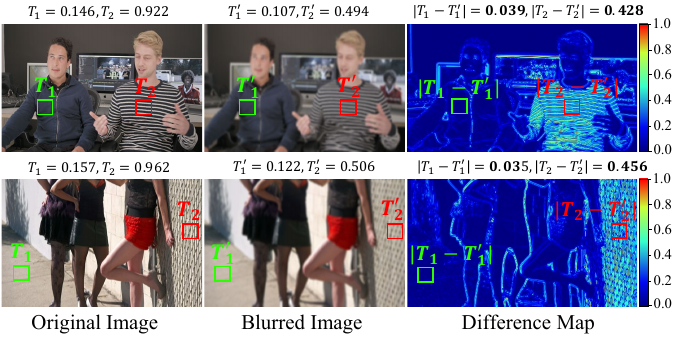}
	}
	\subfigure[Visual comparison of different bin counts] {
		\includegraphics[width=0.40\textwidth]{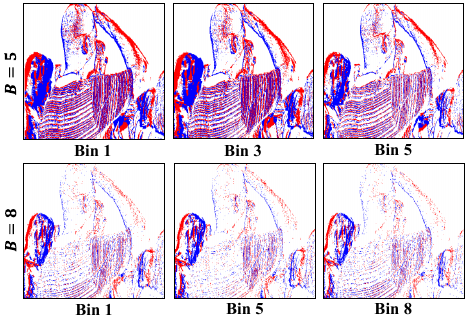}
	}
	\vspace*{-2ex}
	\caption{Discussion of texture magnitude calculation and impact of bin counts. (a) The difference value in high-texture areas is significantly larger than in other areas, indicating that using the contrast of the difference map as a measure of texture magnitude is reasonable. (b) More voxel bin counts lead to weak texture information and the evident noise affecting each voxel bin.}
	\label{fig:figA9}
	\vspace*{-2ex}
\end{figure}

In Fig.~\ref{fig:figA9}(a), we present two examples to demonstrate the feasibility of our texture magnitude calculation in Eq.~(\ref{eq:eq9}). For textures $T_1$ and $T_2$ where $T_2$ is richer, after blurring they become $T_1^{'}$ and $T_2^{'}$. $T_2^{'}$ quality degrades greatly after blurring, whereas $T_1^{'}$ remains relatively unchanged. Thus, the difference {\small $|T_{2} - T_{2}^{'}|$} is far greater than {\small $|T_{1} - T_{1}^{'}|$}. Therefore, the difference map between the original and blurred images reflects texture details. We then calculate the average Root mean square (RMS) contrast over the frames to derive the clip's texture magnitude.

Restoring texture regions is a vital and challenging problem in VSR. So far, there are no metrics to measure a clip's texture magnitude in video analysis. Therefore, we draw insights from image texture analysis study~\cite{cai2022tdpn} and provide an auxiliary method in Eq.~(\ref{eq:eq9}) for analyzing the texture magnitude of clips. This method is easy to use, and the evaluation results verify its feasibility. We hope this can provide a better way to investigate texture-related approaches in the future.

\textbf{Discussion 3: Why is performance at iteration 8 worse than at 5 in Tab.~\ref{table:table5}?}

In our iterative texture enhancement module, the number of iterations $N$ is equivalent to the number of voxel bins $B$. As depicted in Fig.~\ref{fig:figA9}(b), with $B=8$, each bin becomes notably sparse. This sparsity, combined with weak texture information and noticeable noise, leads to a decline in performance, particularly in areas with rich textures. This phenomenon is similar to the concept of over-fitting in machine learning~\cite{mehta2019high}, where additional iterations are not always beneficial and can even degrade performance.

\clearpage

\section{Metrics During Training}
\label{section:metrics}

\begin{figure}[H]
	\vspace*{-2ex}
	\centering
	\subfigure {
		\includegraphics[width=0.46\textwidth]{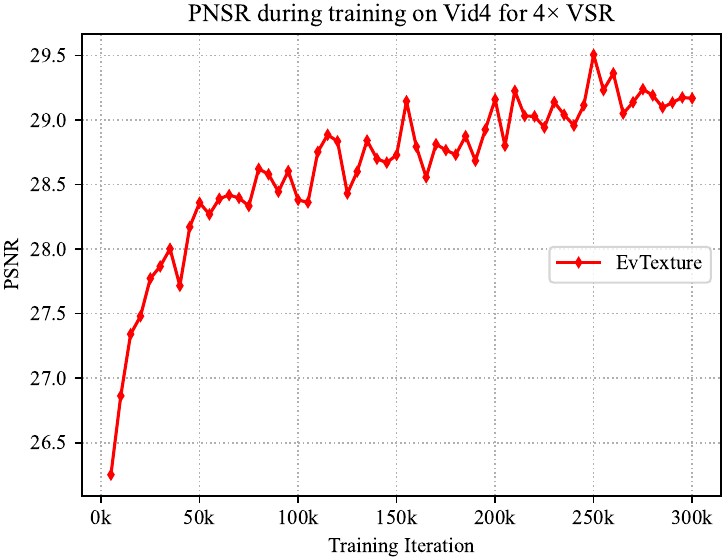}
	}
	\subfigure {
		\includegraphics[width=0.46\textwidth]{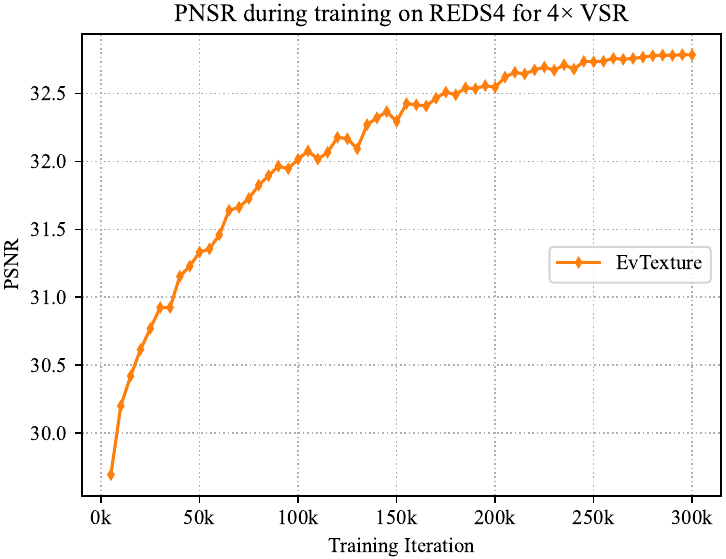}
	}
	\subfigure {
		\includegraphics[width=0.46\textwidth]{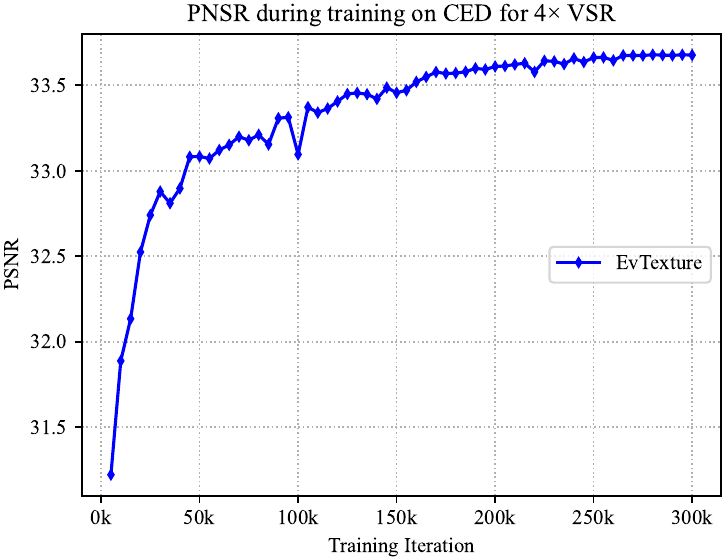}
	}
	\subfigure {
		\includegraphics[width=0.46\textwidth]{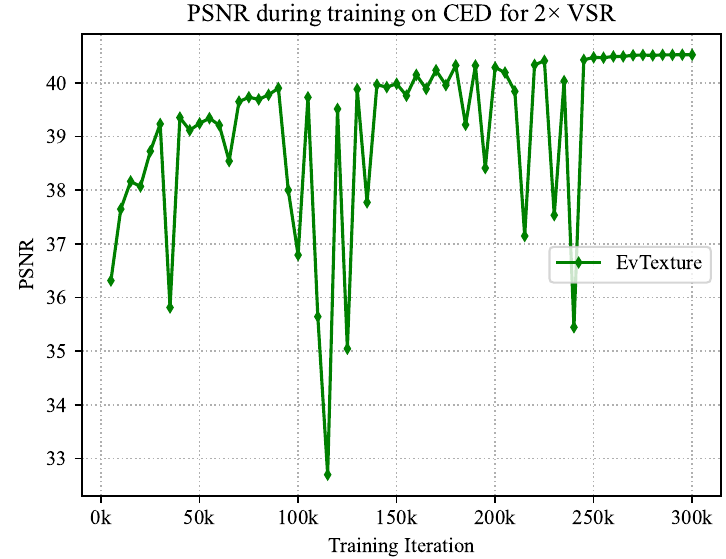}
	}
	\vspace*{-2ex}
	\caption{Metric (PSNR) changes over the iterations of training. These models are all trained from scratch. }
	\label{fig:figA12}
	\vspace*{-2ex}
\end{figure}

Fig.~\ref{fig:figA12} displays plots of PSNR during training, evaluated at saved model checkpoints every 5k iterations over 300k iterations. The curve for 4$\times$ scale training exhibits a gradual convergence. Note that the RGB frames from CED~\cite{scheerlinck2019ced} are created by demosaicing raw frames and suffer from severe noise, also pointed out by Lu~\etal~\yrcite{lu2023learning}. Additionally, since $2\times$ VSR is a relatively easier task, these factors lead to an oscillating validation curve on CED for $2\times$ VSR.

\clearpage

%%%%%%%%%%%%%%%%%%%%%%%%%%%%%%%%%%%%%%%%%%%%%%%%%%%%%%%%%%%%%%%%%%%%%%%%%%%%%%%
%%%%%%%%%%%%%%%%%%%%%%%%%%%%%%%%%%%%%%%%%%%%%%%%%%%%%%%%%%%%%%%%%%%%%%%%%%%%%%%

\end{document}